 \documentclass[pmlr,twocolumn]{jmlr} 
\usepackage{booktabs}

\jmlrvolume{ML4H Proceedings Track}
\jmlryear{2020}
\jmlrsubmitted{2020}
\jmlrworkshop{Machine Learning for Health (ML4H) 2020} 

\usepackage{graphicx}
\usepackage{microtype}
\usepackage{booktabs} 
\usepackage{amsmath}
\usepackage{amsfonts}
\usepackage{verbatim}
\usepackage{varwidth}
\newenvironment{centerverbatim}{%
  \par
  \centering
  \varwidth{\linewidth}
  \verbatim
}{
  \endverbatim
  \endvarwidth
  \par
}
\usepackage{multirow}
\usepackage{mathtools}
\usepackage{algorithm}
\usepackage[noend]{algorithmic}

\SetKwRepeat{Do}{do}{while}

\title[Zero-Shot Clinical Acronym Expansion via Latent Meaning Cells]{Zero-Shot Clinical Acronym Expansion\\ via Latent Meaning Cells}

\author{\Name{Griffin Adams} \Email{griffin.adams@columbia.edu}\\
\Name{Mert Ketenci} \Email{mert.ketenci@columbia.edu}\\
\Name{Shreyas Bhave} \Email{sab2323@cumc.columbia.edu}\\
\Name{Adler Perotte} \Email{adler.perotte@columbia.edu}\\
\Name{No\'emie Elhadad} \Email{noemie.elhadad@columbia.edu}\\
\addr Columbia University, New York, NY, US
}

\begin{document}

\maketitle

\begin{abstract}
We introduce Latent Meaning Cells, a deep latent variable model which learns contextualized representations of words by combining local lexical context and metadata.  Metadata can refer to granular context, such as section type, or to more global context, such as unique document ids. Reliance on metadata for contextualized representation learning is apropos in the clinical domain where text is semi-structured and expresses high variation in topics. We evaluate the LMC model on the task of zero-shot clinical acronym expansion across three datasets. The LMC significantly outperforms a diverse set of baselines at a fraction of the pre-training cost and learns clinically coherent representations.  We demonstrate that not only is metadata itself very helpful for the task, but that the LMC inference algorithm provides an additional large benefit.
\end{abstract}
\section{Introduction}

Pre-trained language models have yielded remarkable advances in multiple natural language processing (NLP) tasks. Probabilistic models such as LDA \citep{blei2003latent}, on the other hand, can uncover latent document-level topics.  In topic models, words are drawn from shared topic distributions at the document level, whereas in language models, word semantics arise from co-occurrence with other words in a tighter window.

We build upon both approaches and introduce Latent Meaning Cells (LMC), a deep latent variable model which learns a contextualized representation of a word by combining evidence from \textit{local context} (i.e., the word and its surrounding words) and \textit{document-level metadata}.  We use the term metadata to generalize the framework because it may vary depending on the domain and application.  Metadata can refer to a document itself, as in topic modeling, document categories (i.e, articles tagged under \textit{Sports}), or structures within documents (i.e., section headers).  Incorporating latent factors into language modeling allows for direct modeling of the inherent uncertainty of words.  As such, we define a latent meaning cell as a Gaussian embedding jointly drawn from word and metadata prior densities. Conditioned on a central word and its metadata, the latent meaning cell identifies surrounding words as in a generative Skip-Gram model \citep{mikolov2013efficient, bravzinskas2017embedding}.  We approximate posterior densities by devising an amortized variational distribution over the latent meaning cells.  The approximate posterior can best be viewed as the embedded word sense based on local context and metadata. In this way, the LMC is non-parametric in the number of latent meanings per word type.

We motivate and develop the LMC model for the task of zero-shot clinical acronym expansion.  Formally, we consider the following task: given clinical text containing an acronym, select the acronym's most likely expansion from a predefined expansion set.  It is analogous to word sense disambiguation, where sense sets are provided by a medical acronym expansion inventory.  This task is important because clinicians frequently use acronyms with diverse meanings across contexts, which makes robust text processing difficult \citep{meyestre2008acronyms, demner2016aspiring}.  Yet clinical texts are highly structured, with established section headers across note types and hospitals \citep{weed1968}. Section headers can serve as a helpful clue in uncovering latent acronym expansions.  For instance, the abbreviation \textit{Ca} is more likely to stand for \textit{calcium} in a \textit{Medications} section whereas it may refer to \textit{cancer} under the \textit{Past Medical History} section.  Prior work has supplemented local word context with document-level features: latent topics \citep{li2019neural} and bag of words \citep{skreta2019training}, rather than section headers.

In our experiments, we directly assess the importance of section headers on zero-shot clinical acronym expansion.  Treating section headers as metadata, we pre-train the LMC model on MIMIC-III clinical notes with extracted sections. Using three test sets, we compare its ability to uncover latent acronym senses to several baselines pre-trained on the same data.  Since labeled data is hard to come by, and clinical acronyms evolve and contain many rare forms \citep{skreta2019training, townsend2013natural}, we focus on the zero-shot scenario: evaluating a model's ability to align the meaning of an acronym in context to the unconditional meaning of its target expansion.  No models are fine-tuned on the task.  We find that metadata complements local word-level context to improve zero-shot performance.  Also, metadata and the LMC model are synergistic - the model's success is a combination of a helpful feature (section headers) and a novel inference procedure.

We summarize our primary contributions: \textbf{(1)} We devise a contextualized language model which jointly reasons over words and metadata.  Previous work has learned document-level representations. In contrast, we explicitly condition the meaning of a word on these representations. \textbf{(2)} Defining metadata as section headers, we evaluate our model on zero-shot clinical acronym expansion and demonstrate superior classification performance.  With relatively few parameters and rapid convergence, the LMC model offers an efficient alternative to more computational intensive models on the task. \textbf{(3)} We publish all code\footnote{\url{https://github.com/griff4692/LMC}} to train, evaluate, and create test data, including regex-based toolkits for reverse substitution and section extraction. This study and use of materials was approved by our institution's IRB.  

\section{Related Work}

\paragraph*{Word Embeddings.} Pre-trained language models learn contextual embeddings through masked, or next, word prediction \citep{peters2018deep, devlin2018bert, yang2019xlnet, bowman2018looking, liu2019roberta, radford2019language}.  Recently, SenseBert \citep{levine2019sensebert} leverages WordNet \citep{miller1998wordnet} to add a masked-word sense prediction task as an auxiliary task in BERT pre-training.  While these models represent words as point embeddings, Bayesian language models treat embeddings as distributions. Word2Gauss defines a normal distribution over words to enable the representation of words as soft regions \citep{vilnis2014word}. Other works directly model polysemy by treating word embeddings as mixtures of Gaussians \citep{tian2014probabilistic, athiwaratkun2017multimodal, athiwaratkun2018probabilistic}. Mixture components correspond to the different word senses. But most of these approaches require setting a fixed number of senses for each word.
Non-parametric Bayesian models enable a variable number of senses per word \citep{neelakantan2015efficient, bartunov2016breaking}. The Multi-Sense Skip Gram model (MSSG) creates new word senses online, while the Adaptive Skip-Gram model \citep{bartunov2016breaking} uses Dirichlet processes. The Bayesian Skip-gram Model (BSG) proposes an alternative to modeling words as a mixture of discrete senses~\citep{bravzinskas2017embedding}.  Instead, the BSG draws latent meaning vectors from center words, which are then used to identify context words.

Embedding models that incorporate global context have also been proposed \citep{le2014distributed, srivastava2013modeling, larochelle2012neural}. The generative models Gaussian LDA, TopicVec, and the Embedded Topic Model (ETM) integrate embeddings into topic models~\citep{blei2003latent}. ETM represents words as categorical distributions with a natural parameter equal to the inner product between word and assigned topic embeddings \citep{dieng2019topic}; Gaussian LDA replaces LDA's categorical topic assumption with multivariate Gaussians \citep{das2015gaussian}; TopicVec can be viewed as a hybrid of LDA and PSDVec \citep{li2016generative}.  While these models make inference regarding the latent topics of a document given words, the LMC model makes inference on meaning given both a word and metadata.

\paragraph*{Clinical Acronym Expansion.}
Acronym expansion---mapping a Short Form (SF) to its most likely Long Form (LF)--- is a task within the problem of word-sense disambiguation \citep{camacho2018word}. 
For instance, the acronym \textit{PT} refers to ``patient'' in \textit{``PT is 80-year old male,''} whereas it refers to ``physical therapy'' in \textit{``prescribed PT for back pain.''}
Traditional supervised approaches to clinical acronym expansion consider only the local context~\citep{joshi2006comparative}. \cite{li2019neural} leverage contextualized ELMo, with attention over topic embeddings, to achieve strong performance after fine-tuning on a randomly sampled MIMIC dataset. On the related task of biomedical entity linking, the LATTE model \citep{zhu2019latte} uses an ELMo-like model to map text to standardized entities in the UMLS meta-thesaurus \citep{bodenreider2004unified}.  \cite{skreta2019training} create a reverse substitution dataset and address class imbalances by sampling additional examples from related UMLS terms.  \cite{jin2019deep} fine-tune bi-ELMO \citep{jin2019probing} with abbreviation-specific classifiers on Pubmed abstracts.

\section{Latent Meaning Cells}

As shown in Figure \ref{fig:lmc}, latent meaning cells postulate both words and metadata as mixtures of latent meanings.

\begin{figure}[htbp]
\floatconts
  {fig:lmc}
  {\caption{The word ``kiwi'' can take on multiple meanings. When used inside a National Geographic article, its latent meaning is restricted to lie inside the red distribution and is closer to ``bird'' than ``fruit''.}}
  {\includegraphics[width=\linewidth]{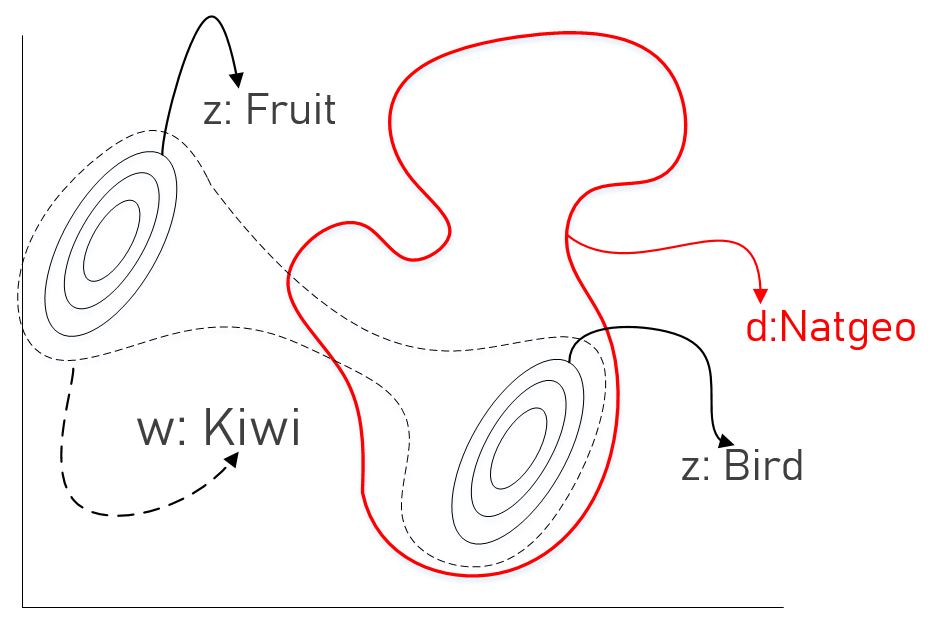}}
\end{figure}

\subsection{Motivation}

In domains where text is semi-structured and expresses high variation in topics, there is an opportunity to consider context between low-level lexical and global document-level.  Clinical texts from the electronic health record represent a prime example.  Metadata, such as section header and note type, can offer vital clues for polysemous words like acronyms. Consequently, we posit that a word's latent meaning directly depends on its metadata.  We define a latent meaning cell (lmc)\footnote{Lowercase \textit{lmc} refers to the latent variable in the uppercase \textit{LMC} graphical model.} as a latent Gaussian embedding jointly drawn from word and metadata prior densities.  The lmc represents a draw of an embedded word sense based on metadata. In a Skip-Gram formulation, we assume that context words are generated from the lmc formed by the center word and corresponding metadata.  Context words, then, are conditionally independent of center words and metadata given the lmc.

\subsection{Notation} \label{notation}

A word is the atomic unit of discrete data and represents an item from a fixed vocabulary.  A word is denoted as $w$ when representing a center word, and $c$ for a context word.  $\boldsymbol{c}$ represents the set of context words relative to a center word $w$.  In different contexts, each word operates as both a center word and a context word.  For our purposes, metadata are pseudo-documents which contain a sequence of $N$ words denoted by $m = (w_1, w_2, ..., w_N)$ where $w_n$ is the $n^{th}$ word. (\ref{metadata-explained} visually depicts metadata). A corpus is a collection of $K$ metadata denoted by $D = \{ m_1, m_2,..., m_K \}$.

\subsection{Latent Variable Setup}

We rely on graphical model notation as a convenient tool for describing the specification of the objective, as is commonly done in latent variable model work (e.g., \citep{bravzinskas2017embedding}).  Using the notation from Section \ref{notation}, we illustrate the pseudo-generative\footnote{We use \textit{pseudo} because the LMC is a latent variable model, not a conventional generative model. As with the Skip-Gram model, due to the re-use of data (center and context words), we cannot use LMC to generate new text, but we can specify an objective function on existing data.} process in plate notation and story form.

\begin{figure}[htbp]
\vskip -0.05in
\label{lmc-box}
\floatconts
    {fig:lmc-box}
    {\includegraphics[width=\columnwidth]{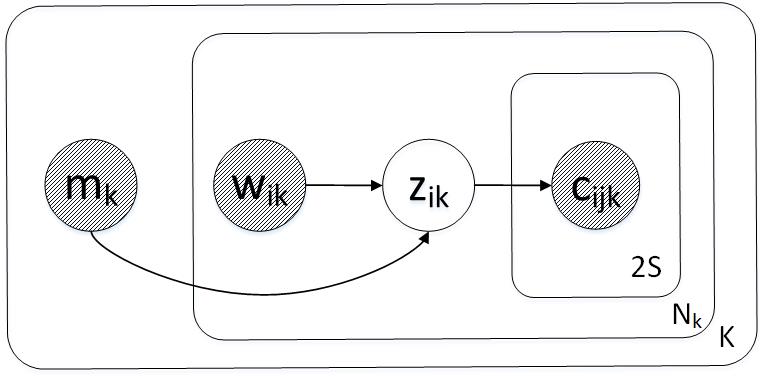}}
    {\caption{LMC Plate Notation.}}
\end{figure}
\vskip -0.15in
\begin{algorithm}
\begin{algorithmic}
\label{alg:gen-story}
\small
\FOR{$k=1...K$}
\STATE Draw metadata $m_{k} \sim \operatorname{Cat}(\gamma)$
\FOR{$i=1...N_k$}
\STATE Draw word $w_{ik} \sim \operatorname{Cat}(\alpha)$
\STATE Draw lmc $z_{ik} \sim p(z_{ik}|w_{ik},m_{k})$
\FOR{$j=1...2S$}
\STATE Draw context word $c_{ijk} \sim p(c_{ijk}|z_{ik})$
\ENDFOR
\ENDFOR
\ENDFOR
\end{algorithmic}
\caption{Pseudo-Generative Story}
\end{algorithm}

$S$ is the window size from which left-right context words are drawn. The factored joint distribution between observed and unobserved random variables $P(M,W,C,Z)$ is:

\vskip -0.15in
\small
$$
\prod_{k=1}^{K}p(m_{k})\prod_{i=1}^{N_{k}}p(w_{ik})p(z_{ik}|w_{ik},m_{k}) \prod_{j=1}^{2S}p(c_{ijk}|z_{ik})
$$
\normalsize

\subsection{Distributions}

We assume the following model distributions: $m_{k} \sim \operatorname{Cat}(\gamma)$, $w_{ik}  \sim \operatorname{Cat}(\alpha)$, and
$z_{ik}|w_{ik},m_{k} \sim N(nn(w_{ik},m_{k};\theta))$. $nn(w_{ik},m_{k};\theta)$ denotes a neural network that outputs isotropic Gaussian parameters.
$p(c_{ijk}|z_{ik})$ is simply a normalized function of fixed parameters ($\theta$) and $z_{ik}$.  We choose a form that resembles Bayes' Rule and compute the ratio of the joint to the marginal:
\begin{equation}
\small
\label{marginal-likelihood}
    p(c_{ijk}|z_{ik}) = \frac{\sum_{m}p(z_{ik}|c_{ijk},m)p(m|c_{ijk})p(c_{ijk})}{\sum_{m}\sum_{c}p(z_{ik}|c,m)p(m|c)p(c)}
\end{equation}
We marginalize over metadata and factorize to include $ p(z_{ik}|c_{ijk},m)$, which shares parameters $\theta$ with $p(z_{ik}|w_{ik},m_{k})$.  The prior over meaning is modeled as in \citet{sohn2015learning}.  $p(m|c)$ and $p(c)$ are defined by corpus statistics. Therefore, the set of parameters that define $p(z_{ik}|w_{ik},m_{k})$ completely determines $p(c_{ijk}|z_{ik})$, making for efficient inference.

\section{Inference} \label{inference}

Ideally, we would like to make posterior inference on lmcs given observed variables.  For one center word $w_{ik}$, this requires modeling
\vskip -0.15in
\small
$$
p(z_{ik}|m_{k},w_{ik},\boldsymbol{c_{ik}}) = \frac{p(z_{ik},m_{k},w_{ik},\boldsymbol{c_{ik}})}{\int p(z_{ik},m_{k},w_{ik},\boldsymbol{c_{ik}})d_{z_{ik}}}
$$
\normalsize

Unfortunately, the posterior is intractable because of the integral. Instead, we use variational Bayes to minimize the KL-Divergence (KLD) between an amortized variational family and the posterior:

\vskip -0.1in
\small
$$
\min_{\phi,\theta} D_{KL}\Big(Q_{\phi}(Z|M,W,C)||P_{\theta}(Z|M,W,C)\Big)
$$
\normalsize
\vskip -0.1in

\subsection{Deriving the Final Objective}

At a high level, we factorize distributions (\ref{factorize-reduce}) and then derive an analytical form of the KLD to arrive at a final objective (\ref{final-objective}).  We then explain the use of approximate bounds for efficiency: the likelihood with negative sampling (\ref{negative-sampling}), and the KLD between the variational distribution and an unbiased mixture estimation (\ref{kl-approx}).

\subsubsection{Final Objective} \label{final-objective}

To avoid high variance, we derive the analytical form of the objective function, rather than optimize with score gradients~\citep{ranganath2014black, schulman2015gradient}. For each center word, the loss function we minimize is:

\vskip -0.3in
\begin{multline}
\label{final-objective-eq}
L_{\phi,\theta}(m_{k},w_{ik},\boldsymbol{c_{ik}}) = \sum_{j=1}^{2S}\max\Bigg(0,\\
D_{KL}\Big(q_{ik}||\sum_{m}p_{\theta}(z_{ik}|c_{ijk},m)\beta_{m|c_{ijk}}\Big)\\
-D_{KL}\Big(q_{ik}||\sum_{m}p_{\theta}(z_{ik}|\Tilde{c},m)\beta_{m|\Tilde{c}}\Big)\Bigg)\\
+D_{KL}\Big(q_{ik}||p_{\theta}(z_{ik}|m_{k},w_{ik})\Big)
\end{multline}
\vskip -0.1in

where $q_{ik}$ denotes $q_{\phi}(z_{ik}|m_{k},w_{ik},\boldsymbol{c_{ik}})$. $\Tilde{c}$ represents a negatively sampled word.  We denote the empirical likelihoods of metadata given a context / negatively sampled word as $\beta_{m|c_{ijk}}$ / $\beta_{m|\Tilde{c}}$. Intuitively, the objective rewards reconstruction of context words through the approximate posterior while encouraging it not to stray too far from the center word's marginal meaning across metadata.  We include the full derivation in \ref{full-derivation}.

\subsubsection{Negative Sampling} \label{negative-sampling}

As in the BSG model, we use negative sampling as an efficient lower bound of the marginal likelihood from Equation \ref{marginal-likelihood}. $\Tilde{c}$ is sampled from the empirical vocabulary distribution $p(\Tilde{c})$ to construct an unbiased estimate for $E_{\Tilde{c}}\Big[\sum_{m}p_{\theta}(z_{ik}|\Tilde{c},m)\beta_{m|\Tilde{c}}\Big]$.  Finally, we transform the likelihood into a hard margin to bound the loss and stabilize training.

\subsubsection{KL-Divergence for Mixtures} \label{kl-approx}

The objective requires computing the KLD between a Gaussian ($q_{ik}$) and a Gaussian mixture ($\sum_{m}p_{\theta}(z|c,m)\beta_{m|c}$). To avoid computing the full marginal, for both context words and negatively sampled words, we sample ten metadata using the appropriate empirical distribution: $\beta_{m|c_{ijk}}$ and $\beta_{m|\Tilde{c}}$, respectively. Using this unbiased sample of mixtures, we form an upper bound for the KLD between the variational family and an unbiased mixture estimation \citep{klmixtures}: $ D_{KL}(f||g) \leq \sum_{a,b}\pi_{a}\omega_{b}D_{KL}(f_{a}||g_{b}) $. $\pi_{a}$ is the mixture weight of $f$ and $\omega_{b}$ is the mixture weight of $g$. $f$ is the variational distribution formed by a single Gaussian and $g$ is the mixture of interest. Thus, the upper-bound is simply the weighted sum of the KLD between the variational distribution and each mixture component.

\subsection{Training Algorithm}

\vskip -0.1in
\begin{algorithm2e}
   \caption{LMC Training Procedure}
   \label{alg:lmc}
Randomly initialize parameters: $\phi,\theta$\
\While{not converged} {
Sample mini-batch $m_{k},w_{ik},\boldsymbol{c_{ik}} \sim D$\
$\delta \xleftarrow{} \nabla_{\phi,\theta} L_{\phi,\theta}(m_{k},w_{ik},\boldsymbol{c_{ik}})$\

$\phi,\theta \xleftarrow{}$ Update using gradient $\delta$\
}
\end{algorithm2e}

The training procedure samples a center word, context word sequence, and metadata from the data distribution and minimizes the loss function from Equation \ref{final-objective-eq} with stochastic gradient descent. In Algorithm \ref{alg:lmc}, we jointly update the variational family and model parameters, $\phi$ and $\theta$ respectively.

\section{Neural Networks}\label{nn}

The LMC model requires modeling two Gaussian distributions, $q_{\phi}(z_{ik}|m_{k},w_{ik},\boldsymbol{c_{ik}})$ and $p_{\theta}(z_{ik}|c_{ijk},m)$.  We parametrize both with neural networks, but any black-box function suffices.  We refer to $q_{\phi}$ as the \texttt{variational network} and $p_{\theta}$ as the \texttt{model network}.

\subsection{Variational Network ($q_{\phi}$)}
The variational network accepts a center word $w_{ik}$, metadata $m_{k}$, and a sequence of context words $\boldsymbol{c_{ik}}$, and outputs isotropic Gaussian parameters: a mean vector $\mu_q$ and variance scalar $\sigma_q$.  Then,  $q_{\phi} \sim N(\mu_q, \sigma_q)$.  At a high level, we encode words with a bi-LSTM \citep{graves2005bidirectional}, summarize the sequence with metadata-specific attention, and then learn a gating function to selectively combine evidence. \ref{variational-model-architecture} contains the full specification.

\subsection{Model Network ($p_{\theta}$)}
The model network accepts a word $w_{ik}$ and metadata $m_{k}$ and projects them onto a higher dimension with embedding matrix $R$. $R_{w_{ik}}$ and $R_{m_{k}}$ are combined: $ h = ReLU(W_{model}([R_{w_{ik}}; R_{m_{k}}]) + b)$. The hidden state $h$ is then separately projected to produce a mean vector $\mu_{p}$ and variance scalar $\sigma_{p}$.  Then, $p_{\theta} \sim N(\mu_p, \sigma_p)$.

\section{Experimental Setup} \label{main-setup}

We pre-train the LMC model and all baselines on unlabeled MIMIC-III notes and compare zero-shot performance on three acronym expansion datasets.  Because we consider the zero-shot scenario, we restrict ourselves to pre-trained contextualized embedding models without fine-tuning.  Out of fidelity to the data, we do not adjust the natural class imbalances.  We explicitly test each model's ability to handle rare expansions, for which shared statistical strength from metadata may be critical.  All models receive the same local word context, yet only two models (MBSGE, LMC) receive section header metadata.  We include full details for Section \ref{main-setup} in \ref{setup}.

\subsection{Pre-Training}
MIMIC-III contains de-identified clinical records from patients admitted to Beth Israel Deaconess Medical Center~\citep{johnson2016mimic}. It comprises two million documents spanning sixteen note types, from discharge summaries to radiology reports. Section headers are extracted through regular expressions.  We pre-train all models for five epochs in PyTorch \citep{paszke2017automatic} and report results on one test set using the others for validation.

\subsection{Evaluation Data}
It is difficult to acquire annotated data for clinical acronym expansion, especially with relevant metadata.  One of the few publicly available datasets with section header annotations is the Clinical Abbreviation Sense Inventory (\textbf{CASI}) dataset \citep{moon2014sense}.  Human annotators assign expansion labels to a set of 74 clinical abbreviations in context.  The authors remove ambiguous examples (based on local word context alone) before publishing the data.  Our experimental test set comprises 27,209 examples across 41 unique acronyms and 150 expansions.

To evaluate across a range of institutions, as well as consider all examples (even ambiguous), we use the acronym sense inventory from CASI to construct two new synthetic datasets via reverse substitution (RS).  RS involves replacing long form expansions with their corresponding short form and then assigning the original expansion as the target label \citep{finley2016towards}.  44,473 tuples of (short form context, section header, target long form) extracted from MIMIC comprise the \textbf{MIMIC RS} dataset.  The second RS dataset consists of 22,163 labeled examples from a corpus of 150k ICU/CCU notes collected between 2005 and 2015 at the Columbia University Irving Medical Center (\textbf{CUIMC}).  For each RS datasest, we draw at most 500 examples per acronym-expansion pair.  For non-MIMIC datasets, when a section does not map to one in MIMIC, we choose the closest corollary.

\subsection{Baselines}

\paragraph*{Dominant \& Random Class.} Acronym expansion datasets are highly imbalanced.  Dominant class accuracy, then, tends to be high and is useful for putting metrics into perspective.  Random performance provides a crude lower bound.

\paragraph*{Section Header MLE.} To isolate the discriminative power of section headers, we include a simple baseline which selects LFs based on $p(LF|section) \propto p(section|LF)$.  We compute $p(section|LF) = \frac{C(section, LF)}{C(LF)}$ on held-out data.

\paragraph*{Bayesian Skip-Gram (BSG).} We implement our own version of the BSG model so that it uses the same \texttt{variational network} architecture as the LMC, with the exception that metadata is unavailable.

\paragraph*{Metadata BSG Ensemble (MBSGE).}  To isolate the added-value of metadata, we devise an ensembled BSG. MBSGE maintains an identical optimization procedure with the exception that it treats metadata and center words as interchangeable observed variables. During training, center words are randomly replaced with metadata, which take on the function of a center word.  For evaluation, we average ensemble the contextualized embeddings from metadata and center word.   We train on two metadata types: section headers and note type, but for experiments, based on available data, we only use headers.

\paragraph*{ELMo.} We use the AllenNLP implementation with default hyperparameters for the Transformer-based version \citep{Gardner2017AllenNLP,peters2018elmotransformer}. We pre-train the model for five epochs with a batch size of 3,072.  We found optimal performance by taking the sequence-wise mean rather than selecting the hidden state from the SF index.

\paragraph*{BERT.} Due to compute limitations, we rely on the publicly available Clinical BioBERT for evaluation \citep{alsentzer2019publicly, lee2020biobert}.  We access the pre-trained model through the Hugging Face Transformer library \citep{wolf2019transformers}.  The weights were initialized from BioBERT (introduces Pubmed articles) before being fine-tuned on the MIMIC-III corpus.  We experimented with many pooling configurations and found that taking the average of the mean and max from the final layer performed best on a validation set.  Another ClinicalBERT uses this configuration \citep{huang2019clinicalbert}.

\subsection{Task Definition}

We rank each candidate acronym expansion (LF) by measuring similarity between its context-independent representation and the contextualized acronym representation.  Table \ref{evaluation-eq-table} shows the ranking functions we used.
$ELMO_{avg}$ represents the mean of final hidden states. For the LMC scoring function, $\sum_{m}p(z|LF_{k},m)\beta_{m|LF_{k}})$ represents the smoothed marginal distribution of a word (or phrase) over metadata (as detailed in \ref{as-vectors}).

\begin{table}[htbp]
\vskip -0.2in
\setlength{\tabcolsep}{-3pt}
\floatconts
  {evaluation-eq-table}%
  {\caption{$LF_{k}$ represents the $k^{th}$ LF.}}%
  {\begin{tabular}{lr}
  \bfseries Model & \bfseries Ranking Function\\
BERT & \scriptsize $ Cosine({BERT_{avg}^{max}(SF; \boldsymbol{c})}, BERT_{avg}^{max}(LF_{k})) $ \\
ELMo & \scriptsize $ Cosine({ELMO_{avg}(SF; \boldsymbol{c})}, ELMO_{avg}(LF_{k})) $ \\
BSG & \scriptsize $ D_{KL}(q(z|SF, \boldsymbol{c}) || p(z|LF_{k})) $ \\
MBSGE & \scriptsize $ D_{KL}(Avg_{x \in\{{SF,m}\}}(q(z|x,\boldsymbol{c})) || p(z|LF_{k})) $ \\
LMC & \scriptsize $ D_{KL}(q(z|SF, m, \boldsymbol{c}) || \sum_{m}p(z|LF_{k},m)\beta_{m|LF_{k}})  $ \\
  \end{tabular}}
 \vskip -0.3in
\end{table}

\section{Results}

\subsection{Classification Performance} \label{classification-results}

\begin{figure*}[ht]
\vskip -0.2in
\begin{center}
\centerline{\includegraphics[width=\columnwidth * 2]{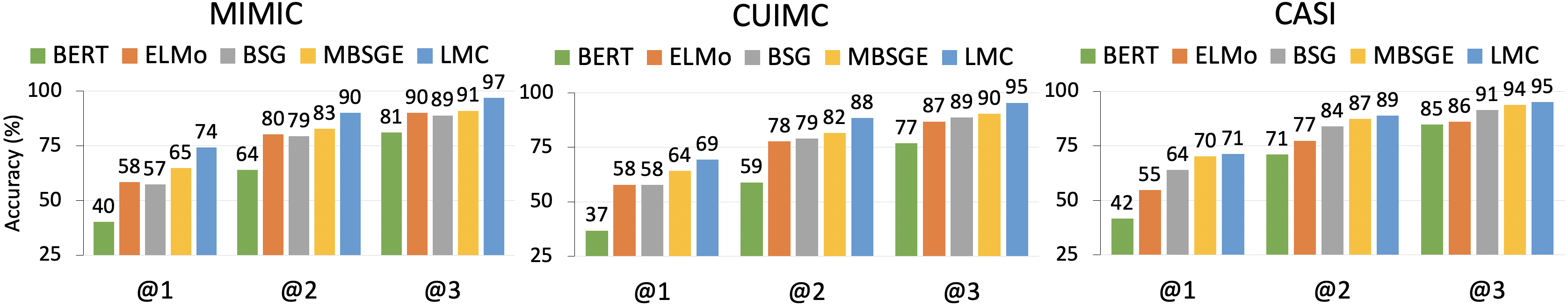}}
\caption{Average accuracy @K across 5 pre-training runs.}
\label{accuracy-at-k}
\end{center}
\vskip -0.4in
\end{figure*}

Recent work has shown that randomness in pre-training contextualized LMs can lead to large variance on downstream tasks \citep{dodge2020fine}.  For robustness, then, we pre-train five separate weights for each model class and report aggregate results.  Tables \ref{acronym-table} and \ref{accuracy-at-k} show mean statistics for each model across five pre-training runs.  In \ref{acronym-results-full}, we show best/worst performance, as well as bootstrap each test set to generate confidence intervals (\ref{bootstrap}).  These additional experiments add robustness and reveal de minimus variance between LMC pre-training runs and between bootstrapped test sets for a single model.  Our main takeaways are:

\begin{table*}[t]
\caption{Mean across 5 pre-training runs. NLL is neg log likelihood, W/M weighted/macro.}
\label{acronym-table}
\begin{center}
\begin{scriptsize}
\begin{sc}
\begin{tabular}{c|cccc|cccc|cccc}
\toprule
\multicolumn{13}{c}{\;\;\;\;\;\;\;\;\;\;\;\;\;\;\;\;\;\textbf{MIMIC}\;\;\;\;\;\;\;\;\;\;\;\;\;\;\;\;\;\;\;\;\;\;\;\;\;\;\;\;\;\;\;\;\;\;\;\;\;\;\textbf{CUIMC}\;\;\;\;\;\;\;\;\;\;\;\;\;\;\;\;\;\;\;\;\;\;\;\;\;\;\;\;\;\;\;\;\;\;\;\;\;\;\;\;\;\textbf{CASI}\;\;}\\
\toprule
Model & NLL & Acc & W F1 & M F1 & NLL & Acc & W F1 & M F1 & NLL & Acc & W F1 & M F1 \\
\midrule
BERT & 1.36 & 0.40 & 0.40 & 0.33 & 1.41 & 0.37 & 0.33 & 0.28 & 1.23 & 0.42 & 0.38 & 0.23\\
ELMo & 1.33 & 0.58 & 0.61 & 0.53 & 1.38 & 0.58 & 0.60 & 0.49 & 1.21 & 0.55 & 0.56 & 0.38\\
BSG & 1.28 & 0.57 & 0.59 & 0.52 & 9.04 & 0.58 & 0.58 & 0.46 & 0.99 & 0.64 & 0.64 & 0.41\\
MBSGE & 1.07 & 0.65 & 0.67 & 0.59 & 6.16 & 0.64 & 0.64 & 0.52 & 0.88 & 0.70 & 0.70 & 0.46\\
LMC & \textbf{0.81} & \textbf{0.74} & \textbf{0.78} & \textbf{0.69} & \textbf{0.90} & \textbf{0.69} & \textbf{0.68} & \textbf{0.57} & \textbf{0.79} & \textbf{0.71} & \textbf{0.73} & \textbf{0.51} \\
\bottomrule
\end{tabular}
\end{sc}
\end{scriptsize}
\end{center}
\end{table*}

\paragraph*{Metadata.} The MBSGE and LMC models materially outperform non-metadata baselines, which suggests that metadata is complementary to local word context for the task.

\paragraph*{LMC Robust Performance.} The LMC outperforms all baselines and exhibits very low variance across pre-training runs.  Given the same input and very similar parameters as MBSGE, the LMC model appears useful beyond the addition of a helpful feature.

\paragraph*{Dataset Comparison.} Unsurprisingly, performance is best on the MIMIC RS dataset because all models are pre-trained on MIMIC notes.  While CUIMC and CASI are in-domain, there is minor performance degradation from the transfer.

\paragraph*{Lower CASI Spread.} The LMC performance gains are less pronounced on the public CASI dataset. CASI was curated to only include examples whose expansions could be unambiguously deduced from local context by humans.  Hence, the relative explanatory power of metadata is likely dampened.

\paragraph*{Poor BERT, ELMo Performance.} BERT / ELMo underperform across datasets. They are optimized to assign high probability to masked or next-word tokens, not to align embedded representations.  For our zero-shot use case, then, they may represent suboptimal pre-training objectives.  Meanwhile, the BSG, MBSGE, and LMC models are trained to align context-dependent representations (\texttt{variational network}) with corresponding context-independent representations (\texttt{model network}). For evaluation, we simply replace context words with candidate LFs.

\paragraph*{Non-Parametric.} Random/dominant accuracy is 27/42\%, 26/47\%, and 31/78\% for MIMIC, CUIMC, and CASI. Section information alone proves very discriminative on MIMIC (85\% accuracy for Section Header MLE), but, given the sparse distribution, it severely overfits. On CASI/CUIMC, the accuracy plummets to 48/46\% and macro F1 to 35/33\%. While relevant, generalization requires distributional header representations.

\subsection{Qualitative Analysis} \label{qualitative}

\subsubsection{Word-Metadata Gating}

Inside the \texttt{variational network}, the network learns a weighted average of metadata and word level representations. We examine instances where more weight is placed on local acronym context vis-a-vis section header, and vice versa.  Table \ref{gating-table} shows that shorter sections with limited topic diversity (e.g., ``Other ICU Medications") are assigned greater relative weight.  The network selectively relies on each source based on relative informativeness.

\begin{table*}
\floatconts
{gating-table}
{\caption{\texttt{variational network} gating function weights.}}
{\begin{tabular}{p{2.75cm} p{9cm} p{1.5cm} }
\textbf{Target Label} & \textbf{Context Window} & \textbf{Section Weight} \\
\toprule
patent ductus \newline arteriosis & \textit{Hospital Course}: echocardiogram showed normal heart structure with \textbf{PDA} hemodynamically significant & 0.12 \\
\midrule
pulmonary artery & \textit{Tricuspid Valve}: physiologic tr pulmonic valve \textbf{PA} physiologic normal pr & 0.21 \\
\midrule
no acute distress & \textit{General Appearance}: well nourished \textbf{NAD} & 0.38 \\
\midrule
morphine sulfate & \textit{Other ICU Medications}: \textbf{MS} $\langle digit \rangle$ pm & 0.46 \\
\bottomrule
\end{tabular}}
\end{table*}

\begin{figure*}[ht]
\begin{center}
\centerline{\includegraphics[width=2\columnwidth]{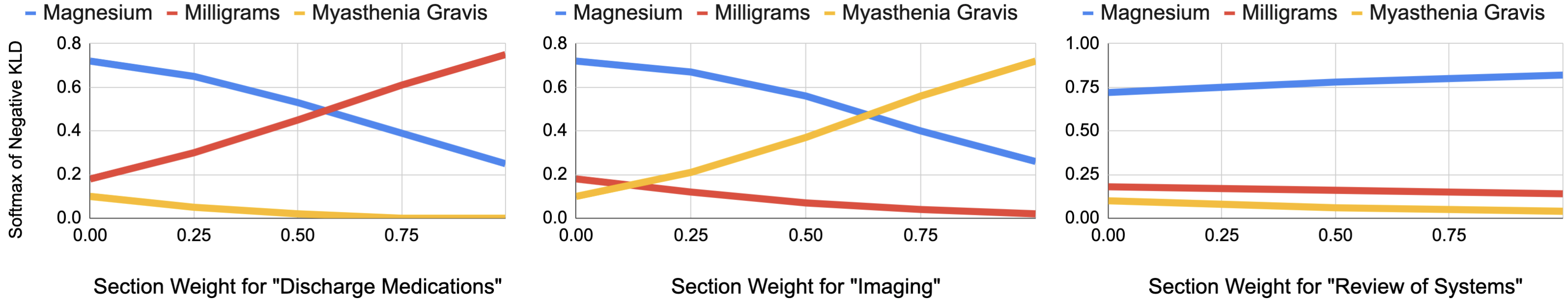}}
\caption{The latent sense distribution changes when manually interpolating the \texttt{variational network} weight between the word ``MG'' \& different section headers. }
\label{interpolation-discharge}
\end{center}
\vskip -0.35in
\end{figure*}

The gating function enables manual interpolation between local context and metadata to measure smoothness in word meaning transitions.  We select three sections which a priori we associate with expansions of the acronym MG: ``Discharge Medications'' with milligrams, ``Imaging'' with myasthenia gravis, and ``Review of Systems'' with magnesium (deficiency). We compute the lmc conditioned on ``MG'' and each section $m$, ranking LFs by taking the softmax over $-D_{KL}(q(z|MG, m, c_{\O}||p(z|LF, m_{\O})))$, where $c_{\O}$ and $m_{\O}$ denote null values. Figure \ref{interpolation-discharge} shows a gradual transition between meanings, suggesting the \texttt{variational network} is a smooth function approximator.

\subsubsection{lmcs as Word Senses}

\begin{table}
\vskip -0.07in
\setlength{\tabcolsep}{-5pt}
\floatconts
{history-table}
{\caption{Conditional latent meaning: history}}
{\begin{tabular}{lr}
\bfseries Section & \bfseries Most Similar Words \\
\toprule
Past Medical History & depression, diabetes \\
Social History & smoking, depression \\
Family History & depression, smoking \\
Glycemic Control & cholesterol, diabetes \\
Left Ventricle & heart, depression \\
Nutrition & diabetes, cholesterol \\
\bottomrule
\end{tabular}}
\vskip -0.15in
\end{table}

A guiding principle behind the LMC model centers on the power of metadata to disambiguate polysemous words.  We choose the word ``history'' and enumerate five diverse types of patient history: smoking, depression, diabetes, cholesterol, and heart.  Then, we examine the proximity of lmcs for the target word under relevant section headers and compare to the expected representations of the five types of patient history.  Section headers have a largely positive impact on word meanings (Table \ref{history-table}), especially for generic words with large prior variances like ``history''.

\subsubsection{Clustering Section Headers}

In Table \ref{cluster-table}, we select five prominent headers and measure cosine proximity of embeddings learned by the \texttt{variational network}\footnote{No difference from using \texttt{model network}.}. In most cases, the results are meaningful, even uncovering a section acronym: ``HPI" for ``History of Present Illness".

\begin{table}
\begin{scriptsize}
\floatconts
{cluster-table}
{\caption{Section header embeddings.}}
{\begin{tabular}{ p{2.25cm} p{4.5cm} }
\toprule
\textbf{Section} & \textbf{Nearest Neighbors} \\
\midrule
Allergies & Social History, Prophylaxis, Disp  \\
\midrule
Chief Complaint & Reason, Family History, Indication \\
\midrule
History of Present Illness & HPI, Past Medical History, \newline Total Time Spent \\
\midrule
Meds on \newline Admission & Discharge Medications, Other Medications, Disp \\
\midrule
Past Medical \newline History & HPI, Social History, \newline History of Present Illness \\
\bottomrule
\end{tabular}}
\vskip -0.2in
\end{scriptsize}
\end{table}

\section{Conclusion}

We target a key problem in clinical text, introduce a helpful feature, and present a Bayesian solution that works well on the task. More generally, the LMC model presents a principled, efficient approach for incorporating metadata into language modeling.

\section{Citations and Bibliography}
\label{sec:cite}

\acks{We thank Arthur Bražinskas, Rajesh Ranganath, and the reviewers for their  constructive, thoughtful feedback. This work was supported by NIGMS award R01 GM114355 and NCATS award U01 TR002062.}

\bibliography{lmc}

\begin{thebibliography}{61}
\providecommand{\natexlab}[1]{#1}
\providecommand{\url}[1]{\texttt{#1}}
\expandafter\ifx\csname urlstyle\endcsname\relax
  \providecommand{\doi}[1]{doi: #1}\else
  \providecommand{\doi}{doi: \begingroup \urlstyle{rm}\Url}\fi

\bibitem[Alsentzer et~al.(2019)Alsentzer, Murphy, Boag, Weng, Jin, Naumann, and
  McDermott]{alsentzer2019publicly}
Emily Alsentzer, John~R Murphy, Willie Boag, Wei-Hung Weng, Di~Jin, Tristan
  Naumann, and Matthew McDermott.
\newblock Publicly available clinical bert embeddings.
\newblock \emph{arXiv preprint arXiv:1904.03323}, 2019.

\bibitem[Athiwaratkun and Wilson(2017)]{athiwaratkun2017multimodal}
Ben Athiwaratkun and Andrew Wilson.
\newblock Multimodal word distributions.
\newblock In \emph{Proceedings of the 55th Annual Meeting of the Association
  for Computational Linguistics (Volume 1: Long Papers)}, pages 1645--1656,
  2017.

\bibitem[Athiwaratkun et~al.(2018)Athiwaratkun, Wilson, and
  Anandkumar]{athiwaratkun2018probabilistic}
Ben Athiwaratkun, Andrew Wilson, and Anima Anandkumar.
\newblock Probabilistic {F}ast{T}ext for multi-sense word embeddings.
\newblock In \emph{Proceedings of the 56th Annual Meeting of the Association
  for Computational Linguistics (Volume 1: Long Papers)}, pages 1--11, 2018.

\bibitem[Bartunov et~al.(2016)Bartunov, Kondrashkin, Osokin, and
  Vetrov]{bartunov2016breaking}
Sergey Bartunov, Dmitry Kondrashkin, Anton Osokin, and Dmitry Vetrov.
\newblock Breaking sticks and ambiguities with adaptive skip-gram.
\newblock In \emph{Artificial Intelligence and Statistics}, pages 130--138,
  2016.

\bibitem[Blei et~al.(2003)Blei, Ng, and Jordan]{blei2003latent}
David~M Blei, Andrew~Y Ng, and Michael~I Jordan.
\newblock Latent {D}irichlet allocation.
\newblock \emph{Journal of Machine Learning Research}, 3:\penalty0 993--1022,
  2003.

\bibitem[Bodenreider(2004)]{bodenreider2004unified}
Olivier Bodenreider.
\newblock The unified medical language system {(UMLS)}: integrating biomedical
  terminology.
\newblock \emph{Nucleic Acids Research}, 32\penalty0 (suppl\_1):\penalty0
  D267--D270, 2004.

\bibitem[Bojanowski et~al.(2017)Bojanowski, Grave, Joulin, and
  Mikolov]{bojanowski2017enriching}
Piotr Bojanowski, Edouard Grave, Armand Joulin, and Tomas Mikolov.
\newblock Enriching word vectors with subword information.
\newblock \emph{Transactions of the Association for Computational Linguistics},
  5:\penalty0 135--146, 2017.

\bibitem[Bowman et~al.(2019)Bowman, Pavlick, Grave, Durme, Wang, Hula, Xia,
  Pappagari, McCoy, Patel, Kim, Tenney, Huang, Yu, Jin, and
  Chen]{bowman2018looking}
Samuel~R. Bowman, Ellie Pavlick, Edouard Grave, Benjamin~Van Durme, Alex Wang,
  Jan Hula, Patrick Xia, Raghavendra Pappagari, R.~Thomas McCoy, Roma Patel,
  Najoung Kim, Ian Tenney, Yinghui Huang, Katherin Yu, Shuning Jin, and Berlin
  Chen.
\newblock Looking for {ELM}o's friends: Sentence-level pretraining beyond
  language modeling, 2019.
\newblock URL \url{https://openreview.net/forum?id=Bkl87h09FX}.

\bibitem[Bra{\v{z}}inskas et~al.(2018)Bra{\v{z}}inskas, Havrylov, and
  Titov]{bravzinskas2017embedding}
Arthur Bra{\v{z}}inskas, Serhii Havrylov, and Ivan Titov.
\newblock Embedding words as distributions with a {B}ayesian skip-gram model.
\newblock In \emph{Proceedings of the 27th International Conference on
  Computational Linguistics}, pages 1775--1789, 2018.

\bibitem[Camacho-Collados and Pilehvar(2018)]{camacho2018word}
Jose Camacho-Collados and Mohammad~Taher Pilehvar.
\newblock From word to sense embeddings: A survey on vector representations of
  meaning.
\newblock \emph{Journal of Artificial Intelligence Research}, 63\penalty0
  (1):\penalty0 743–788, 2018.

\bibitem[Das et~al.(2015)Das, Zaheer, and Dyer]{das2015gaussian}
Rajarshi Das, Manzil Zaheer, and Chris Dyer.
\newblock Gaussian {LDA} for topic models with word embeddings.
\newblock In \emph{Proceedings of the 53rd Annual Meeting of the Association
  for Computational Linguistics and the 7th International Joint Conference on
  Natural Language Processing (Volume 1: Long Papers)}, pages 795--804, 2015.

\bibitem[Demner-Fushman and Elhadad(2016)]{demner2016aspiring}
D~Demner-Fushman and No\'emie Elhadad.
\newblock Aspiring to unintended consequences of natural language processing: a
  review of recent developments in clinical and consumer-generated text
  processing.
\newblock \emph{Yearbook of Medical Informatics}, 25\penalty0 (01):\penalty0
  224--233, 2016.

\bibitem[Devlin et~al.(2019)Devlin, Chang, Lee, and Toutanova]{devlin2018bert}
Jacob Devlin, Ming-Wei Chang, Kenton Lee, and Kristina Toutanova.
\newblock {BERT}: Pre-training of deep bidirectional transformers for language
  understanding.
\newblock In \emph{Proceedings of the 2019 Conference of the North {A}merican
  Chapter of the Association for Computational Linguistics: Human Language
  Technologies, Volume 1 (Long and Short Papers)}, pages 4171--4186, 2019.

\bibitem[Dieng et~al.(2019)Dieng, Ruiz, and Blei]{dieng2019topic}
Adji~B Dieng, Francisco~JR Ruiz, and David~M Blei.
\newblock Topic modeling in embedding spaces.
\newblock \emph{arXiv preprint arXiv:1907.04907}, 2019.

\bibitem[Dodge et~al.(2020)Dodge, Ilharco, Schwartz, Farhadi, Hajishirzi, and
  Smith]{dodge2020fine}
Jesse Dodge, Gabriel Ilharco, Roy Schwartz, Ali Farhadi, Hannaneh Hajishirzi,
  and Noah Smith.
\newblock Fine-tuning pretrained language models: Weight initializations, data
  orders, and early stopping.
\newblock \emph{arXiv preprint arXiv:2002.06305}, 2020.

\bibitem[Finley et~al.(2016)Finley, Pakhomov, McEwan, and
  Melton]{finley2016towards}
Gregory~P Finley, Serguei~VS Pakhomov, Reed McEwan, and Genevieve~B Melton.
\newblock Towards comprehensive clinical abbreviation disambiguation using
  machine-labeled training data.
\newblock In \emph{AMIA Annual Symposium Proceedings}, page 560, 2016.

\bibitem[Gardner et~al.(2018)Gardner, Grus, Neumann, Tafjord, Dasigi, Liu,
  Peters, Schmitz, and Zettlemoyer]{Gardner2017AllenNLP}
Matt Gardner, Joel Grus, Mark Neumann, Oyvind Tafjord, Pradeep Dasigi,
  Nelson~F. Liu, Matthew Peters, Michael Schmitz, and Luke Zettlemoyer.
\newblock {A}llen{NLP}: A deep semantic natural language processing platform.
\newblock In \emph{Proceedings of Workshop for {NLP} Open Source Software
  ({NLP}-{OSS})}, pages 1--6, 2018.

\bibitem[Graves et~al.(2005)Graves, Fern{\'a}ndez, and
  Schmidhuber]{graves2005bidirectional}
Alex Graves, Santiago Fern{\'a}ndez, and J{\"u}rgen Schmidhuber.
\newblock Bidirectional {LSTM} networks for improved phoneme classification and
  recognition.
\newblock In \emph{International Conference on Artificial Neural Networks},
  pages 799--804. Springer, 2005.

\bibitem[Hershey and Olsen(2007)]{klmixtures}
John~R Hershey and Peder~A Olsen.
\newblock Approximating the kullback leibler divergence between gaussian
  mixture models.
\newblock In \emph{2007 IEEE International Conference on Acoustics, Speech and
  Signal Processing-ICASSP'07}, pages IV--317, 2007.

\bibitem[Huang et~al.(2019)Huang, Altosaar, and
  Ranganath]{huang2019clinicalbert}
Kexin Huang, Jaan Altosaar, and Rajesh Ranganath.
\newblock Clinicalbert: Modeling clinical notes and predicting hospital
  readmission.
\newblock \emph{arXiv preprint arXiv:1904.05342}, 2019.

\bibitem[Jin et~al.(2019{\natexlab{a}})Jin, Dhingra, Cohen, and
  Lu]{jin2019probing}
Qiao Jin, Bhuwan Dhingra, William~W Cohen, and Xinghua Lu.
\newblock Probing biomedical embeddings from language models.
\newblock \emph{arXiv preprint arXiv:1904.02181}, 2019{\natexlab{a}}.

\bibitem[Jin et~al.(2019{\natexlab{b}})Jin, Liu, and Lu]{jin2019deep}
Qiao Jin, Jinling Liu, and Xinghua Lu.
\newblock Deep contextualized biomedical abbreviation expansion.
\newblock \emph{arXiv preprint arXiv:1906.03360}, 2019{\natexlab{b}}.

\bibitem[Johnson et~al.(2016)Johnson, Pollard, Shen, Li-wei, Feng, Ghassemi,
  Moody, Szolovits, Celi, and Mark]{johnson2016mimic}
Alistair~EW Johnson, Tom~J Pollard, Lu~Shen, H~Lehman Li-wei, Mengling Feng,
  Mohammad Ghassemi, Benjamin Moody, Peter Szolovits, Leo~Anthony Celi, and
  Roger~G Mark.
\newblock {MIMIC-III}, a freely accessible critical care database.
\newblock \emph{Scientific Data}, 3:\penalty0 160035, 2016.

\bibitem[Joshi et~al.(2006)Joshi, Pakhomov, Pedersen, and
  Chute]{joshi2006comparative}
Mahesh Joshi, Serguei Pakhomov, Ted Pedersen, and Christopher~G Chute.
\newblock A comparative study of supervised learning as applied to acronym
  expansion in clinical reports.
\newblock In \emph{AMIA annual symposium proceedings}, volume 2006, page 399,
  2006.

\bibitem[Kingma and Ba(2014)]{kingma2014adam}
Diederik~P Kingma and Jimmy Ba.
\newblock Adam: A method for stochastic optimization.
\newblock \emph{arXiv preprint arXiv:1412.6980}, 2014.

\bibitem[Larochelle and Lauly(2012)]{larochelle2012neural}
Hugo Larochelle and Stanislas Lauly.
\newblock A neural autoregressive topic model.
\newblock In \emph{Advances in Neural Information Processing Systems}, pages
  2708--2716, 2012.

\bibitem[Le and Mikolov(2014)]{le2014distributed}
Quoc Le and Tomas Mikolov.
\newblock Distributed representations of sentences and documents.
\newblock In \emph{International Conference on Machine Learning}, pages
  1188--1196, 2014.

\bibitem[Lee et~al.(2020)Lee, Yoon, Kim, Kim, Kim, So, and
  Kang]{lee2020biobert}
Jinhyuk Lee, Wonjin Yoon, Sungdong Kim, Donghyeon Kim, Sunkyu Kim, Chan~Ho So,
  and Jaewoo Kang.
\newblock Biobert: a pre-trained biomedical language representation model for
  biomedical text mining.
\newblock \emph{Bioinformatics}, 36\penalty0 (4):\penalty0 1234--1240, 2020.

\bibitem[Levine et~al.(2019)Levine, Lenz, Dagan, Padnos, Sharir,
  Shalev-Shwartz, Shashua, and Shoham]{levine2019sensebert}
Yoav Levine, Barak Lenz, Or~Dagan, Dan Padnos, Or~Sharir, Shai Shalev-Shwartz,
  Amnon Shashua, and Yoav Shoham.
\newblock Sensebert: Driving some sense into bert.
\newblock \emph{arXiv preprint arXiv:1908.05646}, 2019.

\bibitem[Li et~al.(2019)Li, Yasunaga, Nuzumlal{\i}, Caraballo, Mahajan,
  Krumholz, and Radev]{li2019neural}
Irene Li, Michihiro Yasunaga, Muhammed~Yavuz Nuzumlal{\i}, Cesar Caraballo,
  Shiwani Mahajan, Harlan Krumholz, and Dragomir Radev.
\newblock A neural topic-attention model for medical term abbreviation
  disambiguation.
\newblock \emph{arXiv preprint arXiv:1910.14076}, 2019.

\bibitem[Li et~al.(2016)Li, Chua, Zhu, and Miao]{li2016generative}
Shaohua Li, Tat-Seng Chua, Jun Zhu, and Chunyan Miao.
\newblock Generative topic embedding: a continuous representation of documents.
\newblock In \emph{Proceedings of the 54th Annual Meeting of the Association
  for Computational Linguistics (Volume 1: Long Papers)}, pages 666--675, 2016.

\bibitem[Liu et~al.(2019)Liu, Ott, Goyal, Du, Joshi, Chen, Levy, Lewis,
  Zettlemoyer, and Stoyanov]{liu2019roberta}
Yinhan Liu, Myle Ott, Naman Goyal, Jingfei Du, Mandar Joshi, Danqi Chen, Omer
  Levy, Mike Lewis, Luke Zettlemoyer, and Veselin Stoyanov.
\newblock Roberta: A robustly optimized bert pretraining approach.
\newblock \emph{arXiv preprint arXiv:1907.11692}, 2019.

\bibitem[Loper and Bird(2002)]{loper2002nltk}
Edward Loper and Steven Bird.
\newblock {NLTK}: The natural language toolkit.
\newblock In \emph{Proceedings of the {ACL}-02 Workshop on Effective Tools and
  Methodologies for Teaching Natural Language Processing and Computational
  Linguistics}, pages 63--70, 2002.

\bibitem[Meystre et~al.(2008)Meystre, Savova, Kipper-Schuler, and
  Hurdle]{meyestre2008acronyms}
SM~Meystre, GK~Savova, KC~Kipper-Schuler, and JF~Hurdle.
\newblock Extracting information from textual documents in the electronic
  health record: a review of recent research.
\newblock \emph{Yearbook of Medical Informatics}, pages 128--44, 2008.

\bibitem[Mikolov et~al.(2013{\natexlab{a}})Mikolov, Chen, Corrado, and
  Dean]{mikolov2013efficient}
Tomas Mikolov, Kai Chen, Greg Corrado, and Jeffrey Dean.
\newblock Efficient estimation of word representations in vector space.
\newblock \emph{arXiv preprint arXiv:1301.3781}, 2013{\natexlab{a}}.

\bibitem[Mikolov et~al.(2013{\natexlab{b}})Mikolov, Sutskever, Chen, Corrado,
  and Dean]{mikolov2013distributed}
Tomas Mikolov, Ilya Sutskever, Kai Chen, Greg~S Corrado, and Jeff Dean.
\newblock Distributed representations of words and phrases and their
  compositionality.
\newblock In \emph{Advances in Neural Information Processing Systems}, pages
  3111--3119, 2013{\natexlab{b}}.

\bibitem[Miller(1998)]{miller1998wordnet}
George~A Miller.
\newblock \emph{WordNet: An electronic lexical database}.
\newblock MIT press, 1998.

\bibitem[Miyamoto and Cho(2016)]{miyamoto2016gated}
Yasumasa Miyamoto and Kyunghyun Cho.
\newblock Gated word-character recurrent language model.
\newblock In \emph{Proceedings of the 2016 Conference on Empirical Methods in
  Natural Language Processing}, pages 1992--1997, 2016.

\bibitem[Moon et~al.(2014)Moon, Pakhomov, Liu, Ryan, and Melton]{moon2014sense}
Sungrim Moon, Serguei Pakhomov, Nathan Liu, James~O Ryan, and Genevieve~B
  Melton.
\newblock A sense inventory for clinical abbreviations and acronyms created
  using clinical notes and medical dictionary resources.
\newblock \emph{Journal of the American Medical Informatics Association},
  21\penalty0 (2):\penalty0 299--307, 2014.

\bibitem[Neelakantan et~al.(2014)Neelakantan, Shankar, Passos, and
  McCallum]{neelakantan2015efficient}
Arvind Neelakantan, Jeevan Shankar, Alexandre Passos, and Andrew McCallum.
\newblock Efficient non-parametric estimation of multiple embeddings per word
  in vector space.
\newblock In \emph{Proceedings of the 2014 Conference on Empirical Methods in
  Natural Language Processing ({EMNLP})}, pages 1059--1069, 2014.

\bibitem[Paszke et~al.(2017)Paszke, Gross, Chintala, Chanan, Yang, DeVito, Lin,
  Desmaison, Antiga, and Lerer]{paszke2017automatic}
Adam Paszke, Sam Gross, Soumith Chintala, Gregory Chanan, Edward Yang, Zachary
  DeVito, Zeming Lin, Alban Desmaison, Luca Antiga, and Adam Lerer.
\newblock Automatic differentiation in {PyTorch}.
\newblock In \emph{31st Conference on Neural Information Processing Systems
  (NIPS 2017)}, 2017.

\bibitem[Peters et~al.(2018{\natexlab{a}})Peters, Neumann, Iyyer, Gardner,
  Clark, Lee, and Zettlemoyer]{peters2018deep}
Matthew Peters, Mark Neumann, Mohit Iyyer, Matt Gardner, Christopher Clark,
  Kenton Lee, and Luke Zettlemoyer.
\newblock Deep contextualized word representations.
\newblock In \emph{Proceedings of the 2018 Conference of the North {A}merican
  Chapter of the Association for Computational Linguistics: Human Language
  Technologies, Volume 1 (Long Papers)}, pages 2227--2237, 2018{\natexlab{a}}.

\bibitem[Peters et~al.(2018{\natexlab{b}})Peters, Neumann, Zettlemoyer, and
  Yih]{peters2018elmotransformer}
Matthew Peters, Mark Neumann, Luke Zettlemoyer, and Wen-tau Yih.
\newblock Dissecting contextual word embeddings: Architecture and
  representation.
\newblock In \emph{Proceedings of the 2018 Conference on Empirical Methods in
  Natural Language Processing}, pages 1499--1509, 2018{\natexlab{b}}.

\bibitem[Radford et~al.(2019)Radford, Wu, Child, Luan, Amodei, and
  Sutskever]{radford2019language}
Alec Radford, Jeffrey Wu, Rewon Child, David Luan, Dario Amodei, and Ilya
  Sutskever.
\newblock Language models are unsupervised multitask learners.
\newblock \emph{OpenAI Blog}, 1\penalty0 (8):\penalty0 9, 2019.

\bibitem[Ranganath et~al.(2014)Ranganath, Gerrish, and
  Blei]{ranganath2014black}
Rajesh Ranganath, Sean Gerrish, and David Blei.
\newblock Black box variational inference.
\newblock In \emph{Artificial Intelligence and Statistics}, pages 814--822,
  2014.

\bibitem[Schulman et~al.(2015)Schulman, Heess, Weber, and
  Abbeel]{schulman2015gradient}
John Schulman, Nicolas Heess, Theophane Weber, and Pieter Abbeel.
\newblock Gradient estimation using stochastic computation graphs.
\newblock In \emph{Advances in Neural Information Processing Systems}, pages
  3528--3536, 2015.

\bibitem[Skreta et~al.(2019)Skreta, Arbabi, Wang, and
  Brudno]{skreta2019training}
Marta Skreta, Aryan Arbabi, Jixuan Wang, and Michael Brudno.
\newblock Training without training data: Improving the generalizability of
  automated medical abbreviation disambiguation.
\newblock \emph{arXiv preprint arXiv:1912.06174}, 2019.

\bibitem[Sohn et~al.(2015)Sohn, Lee, and Yan]{sohn2015learning}
Kihyuk Sohn, Honglak Lee, and Xinchen Yan.
\newblock Learning structured output representation using deep conditional
  generative models.
\newblock In \emph{Advances in neural information processing systems}, pages
  3483--3491, 2015.

\bibitem[Srivastava et~al.(2013)Srivastava, Salakhutdinov, and
  Hinton]{srivastava2013modeling}
Nitish Srivastava, Ruslan~R Salakhutdinov, and Geoffrey~E Hinton.
\newblock Modeling documents with deep boltzmann machines.
\newblock In \emph{Proceedings of the Twenty-Ninth Conference on Uncertainty in
  Artificial Intelligence (UAI2013)}, 2013.

\bibitem[Srivastava et~al.(2014)Srivastava, Hinton, Krizhevsky, Sutskever, and
  Salakhutdinov]{srivastava2014dropout}
Nitish Srivastava, Geoffrey Hinton, Alex Krizhevsky, Ilya Sutskever, and Ruslan
  Salakhutdinov.
\newblock Dropout: a simple way to prevent neural networks from overfitting.
\newblock \emph{The journal of Machine Learning Research}, 15\penalty0
  (1):\penalty0 1929--1958, 2014.

\bibitem[Strubell et~al.(2019)Strubell, Ganesh, and
  McCallum]{strubell2019energy}
Emma Strubell, Ananya Ganesh, and Andrew McCallum.
\newblock Energy and policy considerations for deep learning in nlp.
\newblock \emph{arXiv preprint arXiv:1906.02243}, 2019.

\bibitem[Tian et~al.(2014)Tian, Dai, Bian, Gao, Zhang, Chen, and
  Liu]{tian2014probabilistic}
Fei Tian, Hanjun Dai, Jiang Bian, Bin Gao, Rui Zhang, Enhong Chen, and Tie-Yan
  Liu.
\newblock A probabilistic model for learning multi-prototype word embeddings.
\newblock In \emph{Proceedings of COLING 2014, the 25th International
  Conference on Computational Linguistics: Technical Papers}, pages 151--160,
  2014.

\bibitem[Townsend(2013)]{townsend2013natural}
Hilary Townsend.
\newblock Natural language processing and clinical outcomes: the promise and
  progress of nlp for improved care.
\newblock \emph{Journal of AHIMA}, 84\penalty0 (2):\penalty0 44--45, 2013.

\bibitem[Vaswani et~al.(2017)Vaswani, Shazeer, Parmar, Uszkoreit, Jones, Gomez,
  Kaiser, and Polosukhin]{vaswani2017attention}
Ashish Vaswani, Noam Shazeer, Niki Parmar, Jakob Uszkoreit, Llion Jones,
  Aidan~N Gomez, {\L}ukasz Kaiser, and Illia Polosukhin.
\newblock Attention is all you need.
\newblock In \emph{Advances in neural information processing systems}, pages
  5998--6008, 2017.

\bibitem[Vilnis and McCallum(2014)]{vilnis2014word}
Luke Vilnis and Andrew McCallum.
\newblock Word representations via gaussian embedding.
\newblock \emph{arXiv preprint arXiv:1412.6623}, 2014.

\bibitem[Vincent et~al.(2008)Vincent, Larochelle, Bengio, and
  Manzagol]{vincent2008extracting}
Pascal Vincent, Hugo Larochelle, Yoshua Bengio, and Pierre-Antoine Manzagol.
\newblock Extracting and composing robust features with denoising autoencoders.
\newblock In \emph{Proceedings of the 25th international conference on Machine
  learning}, pages 1096--1103, 2008.

\bibitem[Weed(1968)]{weed1968}
L~Weed.
\newblock Medical records that guide and teach.
\newblock \emph{New England Journal of Medicine}, 278:\penalty0 593--600, 1968.

\bibitem[Wolf et~al.(2019)Wolf, Debut, Sanh, Chaumond, Delangue, Moi, Cistac,
  Rault, Louf, Funtowicz, et~al.]{wolf2019transformers}
Thomas Wolf, Lysandre Debut, Victor Sanh, Julien Chaumond, Clement Delangue,
  Anthony Moi, Pierric Cistac, Tim Rault, R{\'e}mi Louf, Morgan Funtowicz,
  et~al.
\newblock Transformers: State-of-the-art natural language processing.
\newblock \emph{arXiv preprint arXiv:1910.03771}, 2019.

\bibitem[Yang et~al.(2019)Yang, Dai, Yang, Carbonell, Salakhutdinov, and
  Le]{yang2019xlnet}
Zhilin Yang, Zihang Dai, Yiming Yang, Jaime Carbonell, Russ~R Salakhutdinov,
  and Quoc~V Le.
\newblock {XLN}et: Generalized autoregressive pretraining for language
  understanding.
\newblock In \emph{Advances in neural information processing systems}, pages
  5754--5764, 2019.

\bibitem[Yogatama et~al.(2019)Yogatama, d'Autume, Connor, Kocisky, Chrzanowski,
  Kong, Lazaridou, Ling, Yu, Dyer, et~al.]{yogatama2019learning}
Dani Yogatama, Cyprien de~Masson d'Autume, Jerome Connor, Tomas Kocisky, Mike
  Chrzanowski, Lingpeng Kong, Angeliki Lazaridou, Wang Ling, Lei Yu, Chris
  Dyer, et~al.
\newblock Learning and evaluating general linguistic intelligence.
\newblock \emph{arXiv preprint arXiv:1901.11373}, 2019.

\bibitem[Zhu et~al.(2020)Zhu, Celikkaya, Bhatia, and Reddy]{zhu2019latte}
Ming Zhu, Busra Celikkaya, Parminder Bhatia, and Chandan~K Reddy.
\newblock {LATTE}: Latent type modeling for biomedical entity linking.
\newblock In \emph{AAAI Conference}, 2020.

\end{thebibliography}

\appendix

\onecolumn

\section{Appendix}

\subsection{Future Work}

We hope the LMC framework and code base encourages research into metadata-based language modeling: \textbf{(1) New domains.} The LMC can be applied to any domain which discrete metadata provides informative contextual clues (e.g., document categories, sections, document ids). \textbf{(2) Linguistic Properties.} A unique feature of the LMC is the ability to represent words as marginal distributions over metadata, and vice versa (as detailed in \ref{as-mixtures}).  We encourage exploration into its linguistic implications. \textbf{(3) Metadata Skip-Gram.} Depending on the choice of metadata, the LMC model could be expanded to draw context metadata from a center metadata.  This might capture metadata-level entailment. \textbf{(4) Calibration.} Modeling words and metadata as Gaussian densities can facilitate analysis to connect variance to model uncertainty, instrumental in real-world applications with user feedback. \textbf{(5) Sub-Words.} In morphologically rich languages, subword information has been shown to be highly effective for sharing statistical strength across morphemes \citep{bojanowski2017enriching}. Probabilistic FastText may provide a blueprint for incorporating subwords into LMC \citep{athiwaratkun2018probabilistic}.

\subsection{Metadata Pseudo Document} \label{metadata-explained}

\begin{figure}[h]
    \centering
    \includegraphics[width=\columnwidth]{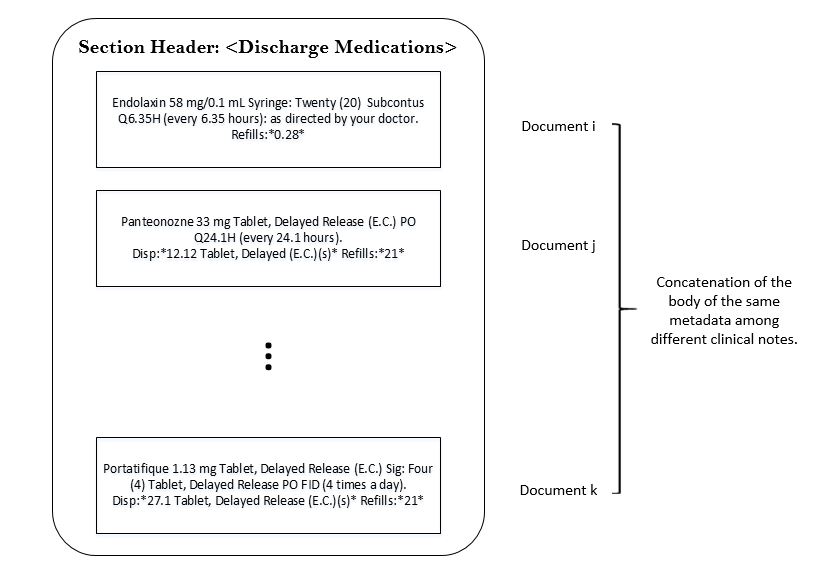}
    \caption{Metadata Pseudo Document for \textit{DISCHARGE MEDICATIONS}.}
    \label{metadata-graphic}
\end{figure}

For our experiments, metadata is comprised of the concatenation of the body of every section header across the corpus.  Yet, when computing context windows, we do not combine text from different physical documents.  Please see Figure \ref{metadata-graphic} for a toy example.

\subsection{Full Derivations} \label{full-derivation}

\subsubsection{Factorize \& Reduce} \label{factorize-reduce}

After factorizing the model posterior and variational distribution, we can push the integral inside the summation and integrate out latent variables that are independent:

\begin{equation}
\sum_{i,k}\int q_{\phi}(z_{ik}|m_{k},w_{ik},\boldsymbol{c_{ik}}) \log{\frac{q_{\phi}(z_{ik}|m_{k},w_{ik},\boldsymbol{c_{ik}})}{p_{\theta}(z_{ik}|m_{k},w_{ik},\boldsymbol{c_{ik}})}} d_{z_{ik}}
\end{equation}
The integral defines a KL measure between individual latent variables, which can be expressed as
\begin{equation}
|W|  \frac{1}{|W|} \sum_{i,k} E_{q_{ik}}\Big[\log{\frac{q_{\phi}(z_{ik}|m_{k},w_{ik},\boldsymbol{c_{ik}})}{p_{\theta}(z_{ik}|m_{k},w_{ik},\boldsymbol{c_{ik}})}}\Big]
\end{equation}
where $|W|$ represents the corpus word count.  Dividing and multiplying by $|W|$ does not change the result:
\begin{equation}
E_{ \hat{p}}\Bigg[D_{KL}\Big(q_{ik}||p_{ik}\Big)\Bigg]
\end{equation}
We ignore $|W|$, as it does not affect the optimization, and denote the amortized variational distribution, model posterior, and the empirical uniform distribution over center words in the corpus as $q_{ik}$, $p_{ik}$, and $\hat{p}$, respectively.

\subsubsection{LMC Objective} \label{full-objective}
In the main manuscript, we outline the steps involved to arrive at the variational objective.  Here, we break it down into a more complete derivation. Because the posterior of the LMC model is intractable, we use variational Bayes and minimize the KLD between the variational distribution and the model posterior:
\begin{equation}
\min D_{KL}\Big(Q(Z|M,W,C)||P(Z|M,W,C)\Big)
\end{equation}
KL-Divergence can also be expressed in expected value form:
\begin{equation}
\min E_{Q}\Big[\log{\frac{Q(Z|M,W,C)}{P(Z|M,W,C)}}\Big]
\end{equation}
The expectation can be re-written in the integral form as follows:
\begin{equation}
\min \int log{\frac{Q(Z|M,W,C)}{P(Z|M,W,C)}}Q(Z|M,W,C) d_{Z}
\end{equation}
Using the independence assumption of the latent random variables, we can factor $Q$ and $P$ as follows:
\begin{equation} 
\min \int ... \int \log{\frac{\prod_{i,k}q(z_{ik}|m_{k},w_{ik},\boldsymbol{c_{ik}})}{\prod_{i,k}p(z_{ik}|m_{k},w_{ik},\boldsymbol{c_{ik}})}} \prod_{i,k}q(z_{ik}|m_{k},w_{ik},\boldsymbol{c_{ik}})d_{z_{ik}}
\end{equation}
Taking the product out of the logarithm yields
\begin{equation}
\min \int ... \int\sum_{i,k}log{\frac{q(z_{ik}|m_{k},w_{ik},\boldsymbol{c_{ik}})}{p(z_{ik}|m_{k},w_{ik},\boldsymbol{c_{ik}})}}\prod_{i,k}q(z_{ik}|m_{k},w_{ik},\boldsymbol{c_{ik}})d_{z_{ik}}
\end{equation}
We can push the integral inside the summation by integrating independent latent variables out:
\begin{equation}
\min \sum_{i,k}\int \log{\frac{q(z_{ik}|m_{k},w_{ik},\boldsymbol{c_{ik}})}{p(z_{ik}|m_{k},w_{ik},\boldsymbol{c_{ik}})}}q(z_{ik}|m_{k},w_{ik},\boldsymbol{c_{ik}})d_{z_{ik}}
\end{equation}
Dividing the summation by the number of words in the corpus defines an expectation over the KL-Divergence for each independent latent variable. Here, $|W|$ denotes the number of words in the corpus. Multiplying the above expression by $|W|$ and dividing by $|W|$ doesn't change the result. Thus,
\begin{equation}
\min  |W|  \frac{1}{|S|} \sum_{i,k} E_{q_{ik}}\Big[\log{\frac{q(z_{ik}|m_{k},w_{ik},\boldsymbol{c_{ik}})}{p(z_{ik}|m_{k},w_{ik},\boldsymbol{c_{ik}})}}\Big]
\end{equation}
$\frac{1}{|W|} \sum_{i,k}$ defines an expectation over the observed data. Therefore, we can write the above expression as
\begin{equation}
\min E_{m_{k},w_{ik},\boldsymbol{c_{ik}} \; \sim \; D }\Bigg[ E_{q_{ik}}\Big[\log{\frac{q(z_{ik}|m_{k},w_{ik},\boldsymbol{c_{ik}})}{p(z_{ik}|m_{k},w_{ik},\boldsymbol{c_{ik}})}}\Big]\Bigg]
\end{equation}
Here the expression $m_{k},w_{ik},\boldsymbol{c_{ik}} \; \sim \; D$ denotes sampling observed variables of document, center word and context words from the data distribution. We ignore $|W|$ as it does not affect the optimization:
\begin{equation}
\min E_{m_{k},w_{ik},\boldsymbol{c_{ik}} \; \sim \; D }\Bigg[D_{KL}\Big(q(z_{ik}|m_{k},w_{ik},\boldsymbol{c_{ik}})||p(z_{ik}|m_{k},w_{ik},\boldsymbol{c_{ik}})\Big)\Bigg]
\end{equation}
The above expression represents the final objective function. To optimize, we sample $m_{k},w_{ik},\boldsymbol{c_{ik}} \; \sim \; D$ and minimize the KL-Divergence between $q$ and $p$. Here $D$ represents the distribution of data from the corpus, which we assume is uniform across observed metadata and words.

\subsubsection{Analytical Form of KL-Divergence} \label{full-kl}
One can approximate KL-Divergence by sampling. Yet, such an estimate has high variance. To avoid this, we derive the analytical form of the objective function.  From Section \ref{full-objective}, we seek to minimize the following objective function:
\begin{equation}
D_{KL}\Big(q_{\phi}(z_{ik}|m_{k},w_{ik},\boldsymbol{c_{ik}})||p_{\theta}(z_{ik}|m_{k},w_{ik},\boldsymbol{c_{ik}})\Big)
\end{equation}
The above equation can be expressed as
\begin{equation}
E_{q_{ik}}\Big[\log{q_{\phi}(z_{ik}|m_{k},w_{ik},\boldsymbol{c_{ik}})} -\log{p_{\theta}(z_{ik},m_{k},w_{ik},\boldsymbol{c_{ik}})}\Big]+\log{p(m_{k},w_{ik},\boldsymbol{c_{ik}})}
\end{equation}
We can factorize $p_{\theta}(z_{ik},m_{k},w_{ik},\boldsymbol{c_{ik}})$ using the model family definition
\begin{equation}
E_{q_{ik}}\Big[\log{q_{\phi}(z_{ik}|m_{k},w_{ik},\boldsymbol{c_{ik}})} -\log{p(m_{k})p(w_{ik})p_{\theta}(z_{ik}|w_{ik},m_{k})\prod_{j=1}^{2S}p_{\theta}(c_{ijk}|z_{ik})}\Big]
+\log{p(m_{k},w_{ik},\boldsymbol{c_{ik}})} 
\label{eq}
\end{equation}
Since, $p(m_{k},w_{ik},\boldsymbol{c_{ik}}) = p(\boldsymbol{c_{ik}}|m_{k},w_{ik})p(w_{ik})p(m_{k})$,
we can re-write Equation \ref{eq} as
\begin{multline}
E_{q_{ik}}\Big[\log{q_{\phi}(z_{ik}|m_{k},w_{ik},\boldsymbol{c_{ik}})} -\log{p(m_{k})} -\log{p(w_{ik})} -\log{p_{\theta}(z_{ik}|w_{ik},m_{k})}\\
-\sum_{j=1}^{2S}\log{p_{\theta}(c_{ijk}|z_{ik})}\Big]+\log{p(\boldsymbol{c_{ik}}|m_{k},w_{ik})+ \log{p(m_{k})} +\log{p(w_{ik})}}
\end{multline}
$\log{p(m_{k})}$ and $\log{p(w_{ik})}$ can leave the expectation and cancel as they do not include any latent variables. Since KL-Divergence is always positive, and the function we are minimizing is the KL-Divergence between the variational family and the posterior, we can write the following inequality:
\begin{equation}
E_{q_{ik}}\Big[\log{q_{\phi}(z_{ik}|m_{k},w_{ik},\boldsymbol{c_{ik}})} -\log{p_{\theta}(z_{ik}|w_{ik},m_{k})}
-\sum_{j=1}^{2S}\log{p_{\theta}(c_{ijk}|z_{ik})}\Big]
+\log{p(\boldsymbol{c_{ik}}|m_{k},w_{ik})} \geq 0
\end{equation}
Pushing the observed variables to the right-hand side of the inequality and negating both sides yields
\begin{equation}
E_{q_{ik}}\Big[-\log{q_{\phi}(z_{ik}|m_{k},w_{ik},\boldsymbol{c_{ik}})} +\log{p_{\theta}(z_{ik}|w_{ik},m_{k})}
+\sum_{j=1}^{2S}\log{p_{\theta}(c_{ijk}|z_{ik})}\Big] \leq \log{p(\boldsymbol{c_{ik}}|m_{k},w_{ik})}
\label{loss-func}\end{equation}
To construct a lower-bound for the likelihood of context words given center word and metadata, $p(\boldsymbol{c_{ik}}|m_{k},w_{ik})$, we minimize the negative left-hand side of Equation \ref{loss-func}. That is, we minimize:
\begin{equation}
\label{elbo}
E_{q_{ik}}\Big[\log{q_{\phi}(z_{ik}|m_{k},w_{ik},\boldsymbol{c_{ik}})} -\log{p_{\theta}(z_{ik}|w_{ik},m_{k})}\Big]
-E_{q_{ik}}\Big[\sum_{j=1}^{2S}\log{p_{\theta}(c_{ijk}|z_{ik})}\Big]
\end{equation}
We can write $E_{q_{ik}}\Big[\log{q_{\phi}(z_{ik}|m_{k},w_{ik},\boldsymbol{c_{ik}})} -\log{p_{\theta}(z_{ik}|w_{ik},m_{k})}\Big]$ as the KL-Divergence between $q_{\phi}(z_{ik}|m_{k},w_{ik},\boldsymbol{c_{ik}})$ and $p_{\theta}(z_{ik}|w_{ik},m_{k})$. That is,
\begin{equation}
D_{KL}\Big(q_{\phi}(z_{ik}|m_{k},w_{ik},\boldsymbol{c_{ik}})|| p_{\theta}(z_{ik}|w_{ik},m_{k})\Big)
-E_{q_{ik}}\Big[\sum_{j=1}^{2S}\log{p_{\theta}(c_{ijk}|z_{ik})}\Big]
\end{equation}
Using the definition of $p(c_{ijk}|z_{ik})$ and re-arranging terms,
\begin{multline}
D_{KL}\Big(q_{\phi}(z_{ik}|m_{k},w_{ik},\boldsymbol{c_{ik}})||p_{\theta}(z_{ik}|w_{ik},m_{k})\Big) \\
-\sum_{j=1}^{2S}E_{q_{ik}}\Big[\log{\sum_{m}p_{\theta}(z_{ik}|c_{ijk},m)p(m|c_{ijk}})p(c_{ijk})\Big]  \\
+E_{q_{ik}}\Big[\log{E_{\Tilde{c}}\Big[\sum_{m}p_{\theta}(z_{ik}|\Tilde{c},m)p(m|\Tilde{c})\Big]}\Big]
\end{multline}
Here, we re-write $\sum_{c}\sum_{d}p_{\theta}(z_{ik}|c,d)p(d|c)p(c)$ in expected value form as $E_{\Tilde{c}}\Big[\sum_{d}p_{\theta}(z_{ik}|\Tilde{c},d)p(d|\Tilde{c})\Big]$. In addition, $p(c_{ijk})$ is the empirical probability value which does not contain the latent variable $z_{ik}$. Therefore, it can leave the expectation and be ignored during optimization:
\begin{multline}
\label{eq-x}
D_{KL}\Big(q_{\phi}(z_{ik}|m_{k},w_{ik},\boldsymbol{c_{ik}})||p_{\theta}(z_{ik}|w_{ik},m_{k})\Big)\\
-\sum_{j=1}^{2S}E_{q_{ik}}\Big[\log{\sum_{m}p_{\theta}(z_{ik}|c_{ijk},m)p(m|c_{ijk}})\Big]\\
+E_{q_{ik}}\Big[\log{E_{\Tilde{c}}\Big[\sum_{m}p_{\theta}(z_{ik}|\Tilde{c},m)p(m|\Tilde{c})\Big]}\Big]
\end{multline}
Adding-subtracting $E_{q_{ik}}[\log{q_{\phi}(z_{ik}|m_{k},w_{ik},\boldsymbol{c_{ik}})}]$ to Equation \ref{eq-x} yields
\begin{multline}
D_{KL}\Big(q_{\phi}(z_{ik}|m_{k},w_{ik},\boldsymbol{c_{ik}})||p_{\theta}(z_{ik}|w_{ik},m_{k})\Big)\\
+\sum_{j=1}^{2S}E_{q_{ik}}\Big[\log{q_{\phi}(z_{ik}|m_{k},w_{ik},\boldsymbol{c_{ik}})}-\log{\sum_{m}p_{\theta}(z_{ik}|c_{ijk},m)p(m|c_{ijk}})\Big]\\
-E_{q_{ik}}\Big[\log{q_{\phi}(z_{ik}|m_{k},w_{ik},\boldsymbol{c_{ik}})}-\log{E_{\Tilde{c}}\Big[\sum_{m}p_{\theta}(z_{ik}|\Tilde{c},m)p(m|\Tilde{c})\Big]}\Big]
\end{multline}
This additional operation defines two KL-Divergence terms:
\begin{multline}
D_{KL}\Big(q_{\phi}(z_{ik}|m_{k},w_{ik},\boldsymbol{c_{ik}})||p_{\theta}(z_{ik}|w_{ik},m_{k})\Big)\\
+\sum_{j=1}^{2S}D_{KL}\Big(q_{\phi}(z_{ik}|m_{k},w_{ik},\boldsymbol{c_{ik}})||\sum_{m}p_{\theta}(z_{ik}|c_{ijk},d)p(m|c_{ijk})\Big)\\
-D_{KL}\Big(q_{\phi}(z_{ik}|m_{k},w_{ik},\boldsymbol{c_{ik}})||{E_{\Tilde{c}}\Big[\sum_{m}p_{\theta}(z_{ik}|\Tilde{c},m)p(m|\Tilde{c})\Big]}\Big)
\end{multline}
To approximate $E_{\Tilde{c}}\Big[\sum_{m}p_{\theta}(z_{ik}|\Tilde{c},m)p(m|\Tilde{c})\Big]$, we sample a word using the negative word distribution (as in word2vec). As in the BSG model, we transform the second term into a hard margin to bound the loss in case the KL-Divergence terms for negatively sampled words are very large.  The final objective we minimize is:

\vskip -0.2in
\begin{multline}
D_{KL}\Big(q_{ik}||p_{\theta}(z_{ik}|m_{k},w_{ik})\Big)+\\
\sum_{j=1}^{2S}\max\Bigg(0,D_{KL}\Big(q_{ik}||\sum_{m}p_{\theta}(z_{ik}|c_{ijk},m)\beta_{m|c_{ijk}}\Big)-D_{KL}\Big(q_{ik}||\sum_{m}p_{\theta}(z_{ik}|\Tilde{c},m)\beta_{m|\Tilde{c}}\Big)\Bigg)
\end{multline}
Here, we denote $q_{\phi}(z_{ik}|m_{k},w_{ik},\boldsymbol{c_{ik}})$ as $q_{ik}$. $\Tilde{c}$ is sampled from $p(c)$ to construct an unbiased estimate for $E_{\Tilde{c}}\Big[\sum_{m}p_{\theta}(z_{ik}|\Tilde{c},m)\beta_{m|\Tilde{c}}\Big]$.

\subsection{\texttt{variational network} Architecture} \label{variational-model-architecture}

Words ($w_{ik}$, $\boldsymbol{c_{ik}}$), as well as metadata ($m_{k}$), are first projected onto a higher dimension via an embedding matrix $E$. The central word embedding $E_{w_{ik}}$ is then tiled across each context word and concatenated with context word embeddings $E_{\boldsymbol{c_{ik}}}$. We then encode the combined word sequence:
\begin{equation}
\boldsymbol{h} = LSTM(\begin{Bmatrix} E_{\boldsymbol{c_{ik}}};  E_{w_{ik}}  \end{Bmatrix})
\end{equation}
where '$;$' denotes concatenation and $\boldsymbol{h}$ represents the concatenation of the hidden states from the forward and backward passes at each timestep. The relevance of a word, especially one with multiple meanings, might depend on the section or document type in which it is found. To allow for an adaptive notion of relevance, we employ scaled dot-product attention \citep{vaswani2017attention} to compute a weighted-average summary of $\boldsymbol{h}$:
\begin{equation}
h_{word} = softmax(\frac{E_{m_{k}}^{T}\boldsymbol{h}}{\sqrt{dim_{e}}})\boldsymbol{h}
\end{equation}
where $dim_{e}$ is the embedding dimension. The scaling factor $\frac{1}{\sqrt{dim_{e}}}$ acts as a normalizer to the dot product.  We selectively combine information from the metadata embedding ($E_{m_{k}}$) and attended context ($h_{word}$) with a gating mechanism similar to \citep{miyamoto2016gated}.  Precisely, we learn a relative weight\footnote{In practice, we compute separate relevance scores for word and metadata and apply the Tanh function before taking the softmax.  We do this to place a constant lower bound on $min(p_{m_{k}}, 1 - p_{m_{k}})$ and prevent over-reliance on one form of evidence.}:
\begin{equation}
p_{m_{k}} = sigmoid(W_{gate}([E_{m_{k}}; h_{word}] + b_{gate}))
\end{equation}
We then use $p_{m_{k}}$ to create a weighted average:
\begin{equation}
h_{joint} = p_{m_{k}} E_{m_{k}} + (1 - p_{m_{k}}) h_{word}
\end{equation}
Finally, we project $h_{joint}$ to produce isotropic Gaussian parameters
\begin{align}
\mu_q = W_{\mu} h_{joint} + b_{\mu}  &&
\sigma_q = exp(W_{\sigma} h_{joint} + b_{\sigma})
\end{align}
As in the BSG model, the network produces the log of the variance, which we exponentiate to ensure it is positive.  We experimented with modeling a full covariance matrix.  Yet, it did not improve performance and added immense cost to the KLD calculation.

\subsection{Additional Details on Experimental Setup}\label{setup}

We provide explanations on a few key design choices for the experimental setup. 

\begin{itemize}
    \item \textbf{MIMIC RS Leakage:} It is important to note that we pre-train all models on the same set of documents which are used to create the synthetic MIMIC RS test set.  While no acronym labels are provided during pre-training, we want to measure, and control for, any train-test leakage that may bias the reporting of the MIMIC RS results.  Yet, we found removing all documents in the test set from pre-training degraded performance no more than one percentage point evenly across all models.  For consistency and computational simplicity, we show performance for models pre-trained on all notes.
    \item \textbf{Mapping section headers from MIMIC to CASI and CUIMC}: We manually map sections in CASI and CUIMC for which no exact match exists in CUIMC.  This is relatively infrequent, and we relied on simple intuition for most mappings.  For example, one such transformation is \textit{Chief Complaint} $\rightarrow$ \textit{Chief Complaints}.
    \item \textbf{Choice of MLE over MAP estimate for section header baseline}: We choose the MLE over MAP estimate because the latter never selects rare LFs due to the huge class imbalances.  This causes macro F1 scores to be very low.
    \item \textbf{LF phrases:} When an LF is a phrase, we take the mean of individual word embeddings.
\end{itemize}

\subsubsection{Preprocessing} \label{appendix-preprocessing}

Clinical text is tokenized, stopwords are removed, and digits are standardized to a common format using the NLTK toolkit \citep{loper2002nltk}. The vocabulary comprises all terms with corpus frequency above 10. We use negative sampling with standard parameter $0.001$ to downsample frequent words \citep{mikolov2013distributed}. After preprocessing, the MIMIC pre-training dataset consists of $\sim330m$ tokens, a token vocabulary size of $\sim100k$, and a section vocabulary size of $\sim10k$.  We write a custom regex to extract section headers from MIMIC notes:
\newline

\begin{centerverbatim}
r'(?:^|\s{4,}|\n)[\d.#]{0,4}\s*([A-Z][A-z0-9/ ]+[A-z]:)'
\end{centerverbatim}

\noindent \newline The search targets a flexible combination of uppercase letters, beginning of line characters, and either a trailing ':' or sufficient space following a candidate header.  We experimented with using template regexes to canonicalize section headers as well as concatenate note type with section headers.  This additional hand-crafted complexity did not improve performance so we use the simpler solution for all experiments.  The code exists to play around with more sophisticated extraction schemes.

\subsubsection{Constructing CASI Test Set} \label{appendix-casi}

For clarity into the results, we outline the filtering operations performed on the CASI dataset.  In Table \ref{filtering-table}, we enumerate the operations and their associated \textcolor{red}{reductions} to the size of the original dataset.  The final dataset at the bottom produces the gold standard test set against which all our models are evaluated.  These changes were made in the interest of producing a coherent test set.  Empirically, performance is not affected by the filtering operations.

\begin{table}[H]
\caption{Filtering CASI Dataset.}
\label{filtering-table}
\begin{center}
\begin{small}
\begin{sc}
\begin{tabular}{lr }
\toprule

Preprocessing Step & Examples \\
\midrule
\midrule
Initial & 37,000 \\
LF Same as SF (just a sense) & \textcolor{red}{5,601} \\
SF Not Present in Context & \textcolor{red}{1,249} \\
Parsing Issue & \textcolor{red}{725} \\
Duplicate Example & \textcolor{red}{731} \\
Single Target & \textcolor{red}{1,481} \\
SFs with LFs not present in MIMIC-III & \textcolor{red}{8,976} \\
\midrule
Final Dataset & \textbf{18,233}  \\
\bottomrule
\end{tabular}
\end{sc}
\end{small}
\end{center}
\end{table}

Because our evaluations rely on computing the distance between contextualized SFs and candidate LFs, we manually curate canonical forms for each LF in the CASI sense inventory.  For instance, we replace the candidate LF for the acronym CVS:
\begin{center}
    "customer, value, service" $\,\to\,$ "CVS pharmacy;brand;store"
\end{center}
where ';' represents a boolean \textit{or}.

\subsubsection{Hyperparameters} \label{hyperparameters}

Our hyperparameter settings are shared across the LMC model and BSG baselines. We assign embedding dimensions of $100d$, and set all hidden state dimensions to $64d$.  We apply a dropout rate of $0.2$ consistently across neural layers \citep{srivastava2014dropout}.  We use a hard margin of $1$ for the hinge loss. Context window sizes are fixed to a minimum of 10 tokens and the nearest section/document boundary.  We develop the model in PyTorch \citep{paszke2017automatic} and train all models for 5 epochs with Adam \citep{kingma2014adam} for adaptive optimization (learning rate of $1e-3$). Inspired by denoising autoencoders \citep{vincent2008extracting} and BERT, we randomly mask context tokens and central words with a probability of 0.2 during training for regularization. The conditional model probabilities $p(w|d)$ and $p(d|w)$ are computed with add-1 smoothing on corpus counts.

\subsubsection{MBSGE Algorithm} \label{mbsge-description}

The training procedure for MBSGE is enumerated in Algorithm \ref{alg:mbsg+}, where
$m_{k}^{1}$ represents the note type for the $k$'th document and $m_{ik}^{2}$ represents the section header corresponding to the $i$'th word in the $k$'th document.  Rather than train three separate models, we train a single model with stochastic replacement to ensure a common embedding space. We choose non-uniform replacement sampling to account for the vastly different vocabulary sizes.
\begin{algorithm}
  \caption{MBSGE Stochastic Training Procedure}
  \label{alg:mbsg+}
\begin{algorithmic}
  \WHILE{not converged}
  \STATE Sample $\boldsymbol{m_k}, w_{ik}, \boldsymbol{c_{ik}} \sim D$
  \STATE Sample $x \sim \operatorname{Cat}(\{w_{ik}, m_{k}^{1}, m_{ik}^{2}\}; \{0.7, 0.1, 0.2\})$
  \STATE $\delta \xleftarrow{} \nabla D_{KL}\Big(q_{\phi}(z_{ik}|x, \boldsymbol{c_{ik}})||p_{\theta}(z_{ik}|x)\Big)$
  \STATE $\phi,\theta \xleftarrow{}$ Update parameters using $\delta$ 
  \ENDWHILE
\end{algorithmic}
\end{algorithm}

For evaluation, we average ensemble the Gaussian parameters from the \texttt{variational network} ($q_\phi)$, where $x$ separately stands for both the center word acronym ($w_{ik}$), and the section header metadata ($m_{ik}^{2}$).

\subsection{Additional Evaluations} \label{additional-evals}

\subsubsection{Aggregate Performance} \label{acronym-results-full}

In the main manuscript, we report mean results across the 5 pre-training runs.  In Table \ref{acronym-table-full}, we include the best and worst performing models to provide a better sense of pre-training variance.  Even though it is a small sample size, it appears the LMC is robust to randomness in weight initialization as evidenced by the tight bounds.

\begin{table*}[t]
\caption{Aggregated across 5 pre-training runs. NLL is neg log likelihood, W/M weighted/macro.}
\label{acronym-table-full}
\begin{center}
\begin{scriptsize}
\begin{tabular}{cc|cccc|cccc|cccc}
\hline
\multicolumn{14}{c}{\;\;\;\;\;\;\;\;\;\;\;\;\;\;\;\;\;\;\;\;\;\;\;\;\;\;\;\;\;\textbf{MIMIC}\;\;\;\;\;\;\;\;\;\;\;\;\;\;\;\;\;\;\;\;\;\;\;\;\;\;\;\;\;\;\;\;\;\;\;\;\;\;\textbf{CUIMC}\;\;\;\;\;\;\;\;\;\;\;\;\;\;\;\;\;\;\;\;\;\;\;\;\;\;\;\;\;\;\;\;\;\;\;\;\;\;\;\;\;\textbf{CASI}}\\
\hline
\toprule
& Model & NLL & Acc & W F1 & M F1 & NLL & Acc & W F1 & M F1 & NLL & Acc & W F1 & M F1 \\
\midrule
\multirow{4}{*}{Worst}
& BERT & 1.36 & 0.40 & 0.40 & 0.33 & 1.41 & 0.37 & 0.33 & 0.28 & 1.23 & 0.42 & 0.38 & 0.23\\
& ELMo & 1.34 & 0.56 & 0.59 & 0.51 & 1.39 & 0.55 & 0.58 & 0.48 & 1.21 & 0.51 & 0.52 & 0.36\\
& BSG & 2.06  & 0.43 & 0.42 & 0.38 & 12.2 & 0.48 & 0.48 & 0.36 & 1.38 & 0.58 & 0.56 & 0.33\\
& MBSGE & 1.26 & 0.60 & 0.62 & 0.54  & 7.94 & 0.61 & 0.61 & 0.48 & 0.96 & 0.68 & 0.67 & 0.43\\
& LMC & \textbf{0.82} & \textbf{0.74} & \textbf{0.77} & \textbf{0.68} & \textbf{0.91} & \textbf{0.69} & \textbf{0.68} & \textbf{0.56} & \textbf{0.80} & \textbf{0.71} & \textbf{0.73} & \textbf{0.50}\\
\midrule
\midrule
\multirow{4}{*}{Mean}
& BERT & 1.36 & 0.40 & 0.40 & 0.33 & 1.41 & 0.37 & 0.33 & 0.28 & 1.23 & 0.42 & 0.38 & 0.23\\
& ELMo & 1.33 & 0.58 & 0.61 & 0.53 & 1.38 & 0.58 & 0.60 & 0.49 & 1.21 & 0.55 & 0.56 & 0.38\\
& BSG & 1.28 & 0.57 & 0.59 & 0.52 & 9.04 & 0.58 & 0.58 & 0.46 & 0.99 & 0.64 & 0.64 & 0.41\\
& MBSGE & 1.07 & 0.65 & 0.67 & 0.59 & 6.16 & 0.64 & 0.64 & 0.52 & 0.88 & 0.70 & 0.70 & 0.46\\
& LMC & \textbf{0.81} & \textbf{0.74} & \textbf{0.78} & \textbf{0.69} & \textbf{0.90} & \textbf{0.69} & \textbf{0.68} & \textbf{0.57} & \textbf{0.79} & \textbf{0.71} & \textbf{0.73} & \textbf{0.51} \\
\midrule
\midrule
\multirow{4}{*}{Best}
& BERT & 1.36 & 0.40 & 0.40 & 0.33 & 1.41 & 0.37 & 0.33 & 0.28 & 1.23 & 0.42 & 0.38 & 0.23\\
& ELMo & 1.33 & 0.61 & 0.65 & 0.58 & 1.38 & 0.62 & 0.64 & 0.50 & 1.21 & 0.59  & 0.60 & 0.42\\
& BSG & 0.98 & 0.64 & 0.68 & 0.59 & 5.41 & 0.61 & 0.62 & 0.50 & 0.85 & 0.67 & 0.70 & 0.46 \\
& MBSGE & 0.96 & 0.68 & 0.71 & 0.62 & 4.81 & 0.67 & 0.67 & 0.57 & 0.83 & \textbf{0.72} & 0.73 & 0.50\\
& LMC & \textbf{0.80} & \textbf{0.75} & \textbf{0.79} & \textbf{0.70} & \textbf{0.89} & \textbf{0.70} & \textbf{0.69} & \textbf{0.58} & \textbf{0.78} & \textbf{0.72} & \textbf{0.74} & \textbf{0.52}\\
\bottomrule
\end{tabular}
\end{scriptsize}
\end{center}
\end{table*}

\subsubsection{Bootstrapping} \label{bootstrap}

\begin{figure}
    \centering
    \includegraphics[width=\textwidth]{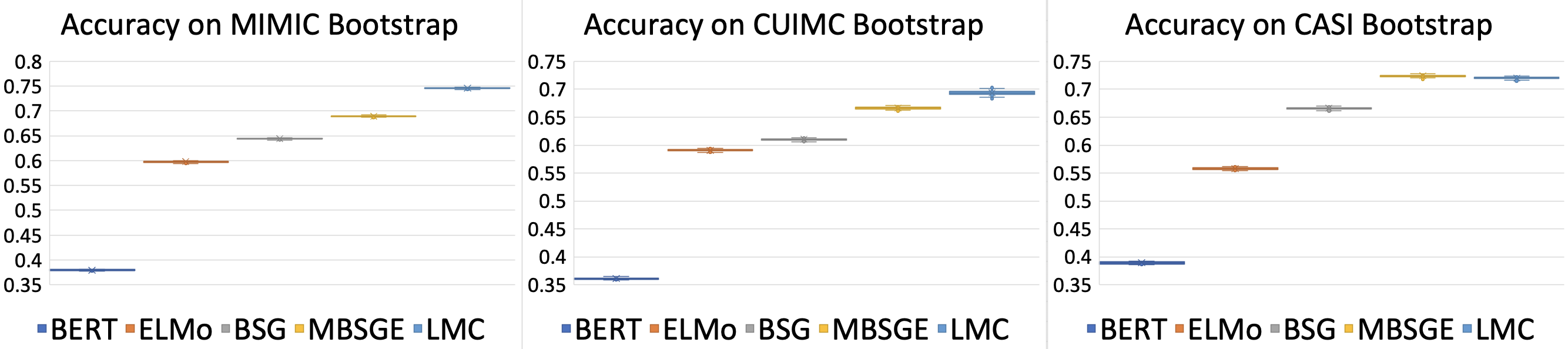}
    \caption{Confidence Intervals for Best Performing Models.}
    \label{confidence-chart}
\end{figure}

For robustness, we select the best performing from each model class and bootstrap the test set to construct confidence intervals. We draw 100 independent random samples from the test set and compute metrics for each model class.  Each subset represents 80\% of the original dataset.  Very tight bounds exist for each model class as can be seen in Figure \ref{confidence-chart}.

\subsubsection{Effect of Number of Target Expansions} \label{performance-by-n}

\begin{figure}[H]
    \includegraphics[width=\textwidth]{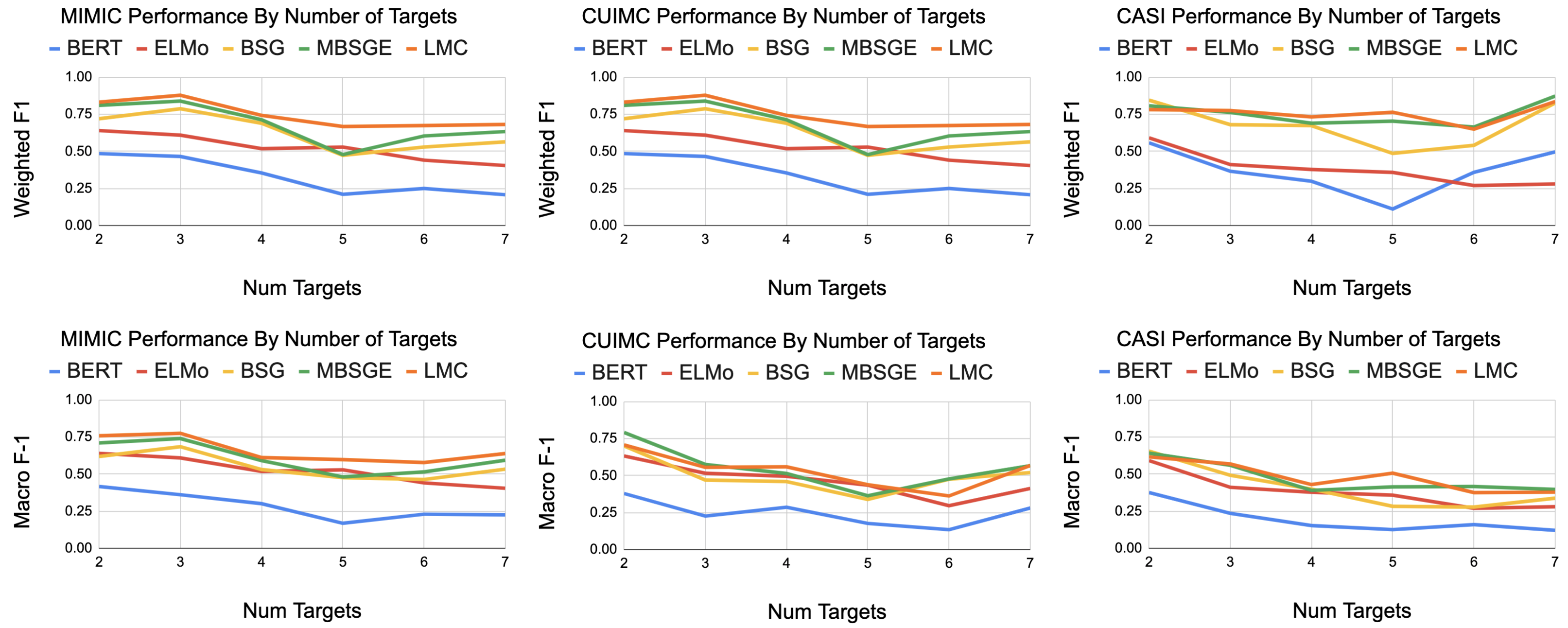}
    \caption{Effect of Number of Output Classes on F1 Performance.  Best performing models shown.}
    \label{perf-by-num-lf}
\end{figure}

For most tasks, performance deteriorates as the number of target outputs grows.  To measure the relative rate of decline, in Figure \ref{perf-by-num-lf}, we plot the F1 score as the number of candidate LFs increases.

\newpage
\subsubsection{Acronym-Level Performance Breakdowns} \label{performance-by-sf}
We provide a breakdown of performance by SF on MIMIC RS between the LMC model and the ELMo baseline.  There is a good deal of volatility across SFs, particularly for the macro F1 metric.  We leave out the other baselines for space considerations.

\begin{small}
\noindent \begin{tabular}{ |p{1.2cm}||p{1.2cm}||p{1.2cm}||p{0.5cm}||p{0.5cm}||p{0.5cm}||p{0.5cm}||p{0.5cm}||p{0.5cm}||p{0.5cm}||p{0.5cm}||p{0.5cm}||p{0.5cm}||p{0.5cm}||p{0.5cm}| }
\hline
\multicolumn{15}{|c|}{$\;\;\;\;\;\;\;\;\;\;\;\;\;\;\;\;\;\;\;\;\;\;\;|\;\;\;\;\;\;\;\;\;\;\;\;\;\;\;\;\;\;\;\;\;\textbf{LMC}$  $\;\;\;\;\;\;\;\;\;\;\;\;\;\;\;\;\;\;\;\;\;\;\;\;\;\;\;\;\;|\;\;\;\;\;\;\;\;\;\;\;\;\;\;\;\;\;\;\;\;\;\textbf{ELMo}$}\\
\hline
Acronym & Count & Targets & mPr & mR & mF1 & wPr & wR & wF1 & mPr & mR & mF1 & wPr & wR & wF1\\
\hline
AMA & 471 & 3 & 0.65 & 0.77 & 0.68 & 0.94 & 0.89 & 0.91 & 0.84 & 0.72 & 0.74 & 0.95 & 0.94 & 0.94\\
ASA & 395 & 2 & 0.5 & 0.5 & 0.5 & 0.98 & 0.99 & 0.99 & 0.5 & 0.5 & 0.5 & 0.98 & 0.99 & 0.99\\
AV & 491 & 3 & 0.57 & 0.69 & 0.58 & 0.88 & 0.79 & 0.82 & 0.58 & 0.41 & 0.13 & 0.92 & 0.08 & 0.11\\
BAL & 485 & 2 & 0.68 & 0.84 & 0.72 & 0.93 & 0.87 & 0.89 & 0.64 & 0.87 & 0.65 & 0.93 & 0.78 & 0.83 \\
BM & 488 & 3 & 0.71 & 0.67 & 0.52 & 0.95 & 0.73 & 0.8 & 0.84 & 0.52 & 0.55 & 0.93 & 0.93 & 0.92 \\ 
CnS & 432 & 5 & 0.53 & 0.67 & 0.56 & 0.96 & 0.96 & 0.96 & 0.63 & 0.67 & 0.41 & 0.99 & 0.18 & 0.26 \\
CEA & 497 & 4 & 0.31 & 0.28 & 0.2 & 0.92 & 0.34 & 0.43 & 0.45 & 0.35 & 0.16 & 0.97 & 0.18 & 0.3 \\ 
CR & 499 & 6 & 0.47 & 0.61 & 0.38 & 0.97 & 0.84 & 0.88 & 0.17 & 0.17 & 0.01 & 0.91 & 0.04 & 0.01 \\ 
CTA & 495 & 4 & 0.49 & 0.44 & 0.46 & 0.98 & 0.94 & 0.96 & 0.51 & 0.89 & 0.49 & 0.97 & 0.85 & 0.91 \\
CVA & 474 & 2 & 0.93 & 0.91 & 0.91 & 0.92 & 0.92 & 0.91 & 0.78 & 0.5 & 0.37 & 0.76 & 0.57 & 0.42 \\ 
CVP & 487 & 3 & 0.61 & 0.76 & 0.51 & 0.92 & 0.63 & 0.75 & 0.45 & 0.56 & 0.44 & 0.91 & 0.72 & 0.77 \\
CVS & 237 & 3 & 0.47 & 0.78 & 0.38 & 0.88 & 0.47 & 0.53 & 0.47 & 0.34 & 0.34 & 0.78 & 0.78 & 0.74 \\
DC & 455 & 5 & 0.53 & 0.72 & 0.51 & 0.74 & 0.55 & 0.62 & 0.17 & 0.29 & 0.2 & 0.43 & 0.56 & 0.48 \\ 
DIP & 492 & 3 & 0.85 & 0.97 & 0.89 & 0.97 & 0.94 & 0.95 & 0.37 & 0.42 & 0.2 & 0.93 & 0.33 & 0.41 \\ 
DM & 484 & 3 & 0.61 & 0.86 & 0.57 & 0.92 & 0.78 & 0.83 & 0.65 & 0.75 & 0.51 & 0.97 & 0.64 & 0.77 \\ 
DT & 475 & 6 & 0.35 & 0.28 & 0.31 & 0.68 & 0.49 & 0.57 & 0.11 & 0.55 & 0.12 & 0.14 & 0.16 & 0.15 \\ 
EC & 473 & 4 & 0.59 & 0.74 & 0.54 & 0.95 & 0.93 & 0.93 & 0.27 & 0.59 & 0.16 & 0.1 & 0.04 & 0.05 \\ 
ER & 495 & 3 & 0.67 & 0.72 & 0.68 & 0.93 & 0.89 & 0.91 & 0.35 & 0.34 & 0.03 & 0.9 & 0.05 & 0.02 \\ 
FSH & 265 & 2 & 0.75 & 0.66 & 0.7 & 0.99 & 0.99 & 0.99 & 0.49 & 0.5 & 0.5 & 0.98 & 0.99 & 0.98 \\ 
IA & 171 & 2 & 0.51 & 0.74 & 0.35 & 0.99 & 0.49 & 0.64 & 0.51 & 0.5 & 0.02 & 0.99 & 0.02 & 0.01 \\ 
IM & 492 & 2 & 0.66 & 0.9 & 0.7 & 0.95 & 0.84 & 0.88 & 0.54 & 0.54 & 0.16 & 0.93 & 0.16 & 0.16 \\ 
LA & 454 & 3 & 0.7 & 0.98 & 0.75 & 0.99 & 0.98 & 0.99 & 0.48 & 0.62 & 0.19 & 0.96 & 0.06 & 0.05 \\ 
LE & 481 & 7 & 0.39 & 0.49 & 0.38 & 0.93 & 0.78 & 0.84 & 0.28 & 0.56 & 0.26 & 0.78 & 0.42 & 0.53 \\ 
MR & 492 & 5 & 0.44 & 0.62 & 0.35 & 0.96 & 0.5 & 0.63 & 0.42 & 0.72 & 0.26 & 0.92 & 0.34 & 0.31 \\ 
MS & 488 & 6 & 0.48 & 0.6 & 0.33 & 0.92 & 0.33 & 0.46 & 0.41 & 0.55 & 0.31 & 0.75 & 0.42 & 0.37 \\ 
NAD & 465 & 2 & 0.4 & 0.5 & 0.44 & 0.64 & 0.8 & 0.71 & 0.58 & 0.54 & 0.54 & 0.72 & 0.77 & 0.73 \\ 
NP & 463 & 4 & 0.44 & 0.58 & 0.48 & 0.93 & 0.87 & 0.89 & 0.53 & 0.38 & 0.32 & 0.91 & 0.88 & 0.84 \\ 
OP & 489 & 6 & 0.59 & 0.57 & 0.57 & 0.91 & 0.91 & 0.9 & 0.52 & 0.66 & 0.57 & 0.78 & 0.85 & 0.81 \\ 
PA & 412 & 6 & 0.38 & 0.48 & 0.29 & 0.82 & 0.46 & 0.44 & 0.43 & 0.43 & 0.26 & 0.92 & 0.35 & 0.36 \\ 
PCP & 488 & 4 & 0.44 & 0.59 & 0.32 & 0.67 & 0.43 & 0.45 & 0.48 & 0.41 & 0.35 & 0.77 & 0.44 & 0.45 \\
PDA & 478 & 3 & 0.49 & 0.53 & 0.48 & 0.83 & 0.74 & 0.75 & 0.86 & 0.8 & 0.82 & 0.8 & 0.81 & 0.79 \\ 
PM & 375 & 3 & 0.4 & 0.38 & 0.21 & 0.83 & 0.32 & 0.28 & 0.46 & 0.4 & 0.41 & 0.78 & 0.81 & 0.78 \\ 
PR & 241 & 4 & 0.67 & 0.75 & 0.58 & 0.82 & 0.71 & 0.76 & 0.6 & 0.35 & 0.31 & 0.78 & 0.47 & 0.39 \\
PT & 496 & 4 & 0.53 & 0.69 & 0.58 & 0.96 & 0.93 & 0.94 & 0.42 & 0.35 & 0.17 & 0.94 & 0.11 & 0.13 \\ 
RA & 490 & 4 & 0.5 & 0.57 & 0.47 & 0.91 & 0.72 & 0.78 & 0.66 & 0.56 & 0.58 & 0.91 & 0.9 & 0.9 \\ 
RT & 470 & 4 & 0.55 & 0.47 & 0.41 & 0.91 & 0.66 & 0.69 & 0.55 & 0.46 & 0.37 & 0.82 & 0.66 & 0.58 \\ 
SA & 454 & 5 & 0.8 & 0.77 & 0.61 & 0.99 & 0.84 & 0.85 & 0.58 & 0.65 & 0.48 & 0.73 & 0.82 & 0.77 \\ 
SBP & 489 & 2 & 0.59 & 0.55 & 0.24 & 0.86 & 0.25 & 0.2 & 0.64 & 0.74 & 0.54 & 0.87 & 0.57 & 0.61 \\ 
US & 290 & 2 & 0.91 & 0.91 & 0.91 & 0.92 & 0.92 & 0.92 & 0.86 & 0.61 & 0.6 & 0.82 & 0.75 & 0.69 \\ 
VAD & 482 & 4 & 0.44 & 0.48 & 0.25 & 0.9 & 0.37 & 0.51 & 0.25 & 0.27 & 0.04 & 0.8 & 0.07 & 0.13 \\ 
VBG & 483 & 2 & 0.79 & 0.75 & 0.7 & 0.83 & 0.71 & 0.7 & 0.96 & 0.95 & 0.95 & 0.96 & 0.95 & 0.95 \\
\bottomrule
$\boldsymbol{AVG}$ & - & - & \textbf{0.57} & \textbf{0.65} & \textbf{0.51} & \textbf{0.9} & \textbf{0.72} & \textbf{0.75} & \textbf{0.52} & \textbf{0.54} & \textbf{0.37} & \textbf{0.83} & \textbf{0.52} & \textbf{0.52} \\
\hline
\end{tabular}
\end{small}

\subsection{Efficiency} \label{efficiency}

\begin{figure}[htbp]
\begin{center}
\floatconts
{efficiency-chart}
{\includegraphics[width=\linewidth]{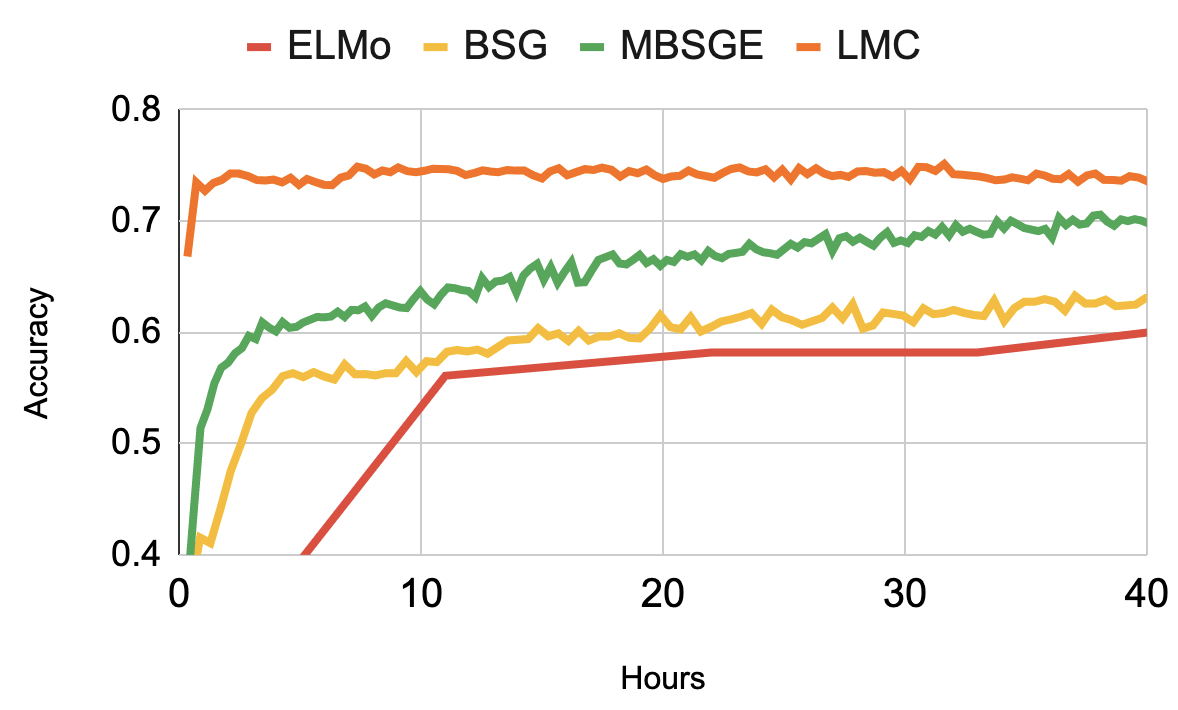}}
{\caption{Accuracy by pre-training hours.  All plots flatten after 40 hours (not shown).}}
\end{center}
\end{figure}

Task performance at the end of pre-training is an informative, but potentially incomplete, evaluation metric.  Recent work has noted that large-scale transfer learning can come at a notable financial and environmental cost \citep{strubell2019energy}. Also, a model which adapts quickly to a task may emulate general linguistic intelligence \citep{yogatama2019learning}.  In Figure \ref{efficiency-chart}, we plot test set accuracy on MIMIC RS at successive pre-training checkpoints.  We pre-train the models on a single NVIDIA GeForce RTX 2080 Ti GPU.  We hypothesize that flexibility in latent word senses and shared statistical strength across section headers facilitate rapid LMC convergence.  Averaged across datasets and runs, the number of pre-training hours required for peak test set performance is 6 for LMC, while 50, 51, and 55 for MBSGE, BSG, and ELMo.  The non-embedding parameter counts are 169k for the LMC and 150k for both the BSG and MBSGE. ELMo has 91mn parameters.  Taken together, the LMC efficiently learns the task as a by-product of pre-training.

\subsection{Words and Metadata as Mixtures}\label{as-mixtures}

Consider metadata and its building blocks. A natural question to consider is the distribution of latent meanings given metadata. We can simply write this as
\begin{equation}
p(z_{ik}|m_{k}) = \sum_{w_{ik}}p(z_{ik}|w_{ik},m_{k})p(w_{ik}|m_{k})
\end{equation}
$w_{ik}$ denotes an arbitrary word in document $k$ and the summation marginalizes it with respect to the vocabulary.
$p(w_{ik}|m_{k})$ can be measured empirically with corpus statistics. We will denote this probability value as $\xi_{w_{ik}|m_{k}}$. In addition, $p(z_{ik}|,w_{ik},m_{k})$ has already been defined as $N(nn(w_{ik},m_{k};\theta))$. Therefore,
\begin{equation}
p(z_{ik}|m_{k}) = \sum_{w_{ik}}N(nn(w_{ik},m_{k};\theta))\xi_{w_{ik}|m_{k}}
\end{equation}
The distribution of the latent space over metadata is a mixture of Gaussians weighted by occurrence probability in metadata $k$. One can measure the similarity between two metadata using KL-Divergence. This measure is computationally expensive because each metadata can be a mixture of thousands of Gaussians. Monte Carlo sampling, however, can serve as an efficient, unbiased approximation.

\begin{figure}[H]
    \centering
    \includegraphics[width=0.75\columnwidth,scale=1]{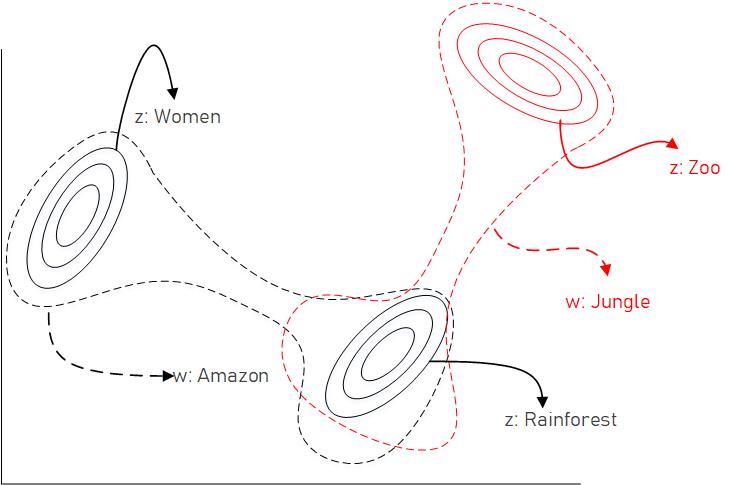}
    \caption{The meaning of ``Amazon'' can be interpreted as a mixture of Gaussian distributions in different metadata.}
    \label{wordmixtures}
    \centering
\end{figure}

It is also a natural question to ask about the potential meanings a word can exhibit (Figure \ref{wordmixtures}). That is,

\begin{equation}
p(z_{ik}|w_{ik}) = \sum_{m_{k}}p(z|m_{k},w_{ik})p(m_{k}|w_{ik})
\end{equation}
$p(m_{k}|w_{ik})$ can also be measured empirically. We denote this distribution as $\beta_{m_{k}|w_{ik}}$.
\begin{equation}
\label{equation:marginal-words}
p(z_{ik}|w_{ik}) = \sum_{m_{k}}N(nn(w_{ik},m_{k};\theta))\beta_{m_{k}|w_{ik}}
\end{equation}

\subsection{Word and Metadata as Vectors}\label{as-vectors}
With a certain trade-off of compression, we can represent metadata as a vector using its expected conditional meaning:
\begin{equation}
E_{z_{ik}|m_{k}}[z_{ik}] = \sum_{w_{ik}}\xi_{w_{ik}|m_{k}}\int z_{ik}N(nn(w_{ik},m_{k};\theta))dz_{ik}
\end{equation}
Since $\int z_{ik}N(nn(w_{ik},m_{k};\theta))dz_{ik} = E_{z_{ik}|w_{ik},m_{k}}[z_{ik}]$ 
The expectation can be simply written as the combination of the means of normal distributions that form metadata $k$:
\begin{equation}
E_{z_{ik}|m_{k}}[z_{ik}] = \sum_{w_{ik}}\xi_{w_{ik}|m_{k}} E[z_{ik}|w_{ik},m_{k}]
\end{equation}
The above equation sums the expected meaning of words inside a metadata weighted by occurrence probability. Following the same logic for words yields
\begin{equation}
\label{words-as-vectors-eq}
E_{z_{ik}|w_{ik}}[z_{ik}] = \sum_{m_{k}}\beta_{m_{k}|w_{ik}}E[z_{ik}|w_{ik},m_{k}]
\end{equation}

\end{document}